\documentclass{article}

\usepackage{booktabs}
\usepackage{wrapfig}
\usepackage{multirow}
\usepackage{hhline}
\usepackage{colortbl}
\usepackage{microtype}
\usepackage{graphicx}
\usepackage{subcaption}

\usepackage{anyfontsize}

\usepackage{booktabs} 

\usepackage{hyperref}
\usepackage{booktabs} 
\usepackage{amsmath,amssymb,amsthm}
\usepackage{boxedminipage}
\usepackage{stmaryrd}
\usepackage{marvosym}
\usepackage{lmodern}
\usepackage{tikz}

\usepackage{cuted}

\usepackage{pgfplots}
\usepackage{pgfplotstable}
\usepackage{xcolor}
\usepackage{tabu}

\usepackage[none]{hyphenat}

\newcommand{\figref}[1]{Fig.~\ref{Fi:#1}}
\newcommand{\tabref}[1]{Table~\ref{Ta:#1}}
\newcommand{\secref}[1]{Section~\ref{Se:#1}}

\usetikzlibrary{quotes,angles,positioning,calc, shapes, arrows, 3d}

\definecolor{darkred}{rgb}{0.5,0,0}
\definecolor{darkgreen}{rgb}{0,0.5,0}
\definecolor{darkblue}{rgb}{0,0,0.5}
\definecolor{lightgrey}{rgb}{0.7,0.7,0.7}

\definecolor{lightergrey}{rgb}{0.93,0.93,0.93}
\newcommand{\clinelight}[1]{\tabucline[lightgrey]{#1}}

\usepackage{placeins}
\usepackage{fancyvrb}

\usepackage[arxiv]{icml2019}

\theoremstyle{definition}
\newtheorem{definition}{Definition}[section]

\icmltitlerunning{A Provable Defense for Deep Residual Networks}

\begin{document}

\setlength{\abovedisplayskip}{3pt}
\setlength{\belowdisplayskip}{3pt}

\twocolumn[
\icmltitle{A Provable Defense for Deep Residual Networks}

\begin{icmlauthorlist}
\icmlauthor{Matthew Mirman}{eth}
\icmlauthor{Gagandeep Singh}{eth}
\icmlauthor{Martin Vechev}{eth}
\end{icmlauthorlist}

\icmlaffiliation{eth}{Department of Computer Science, ETH Zurich, Zurich, Switzerland}

\icmlcorrespondingauthor{Matthew Mirman}{matthew.mirman@inf.ethz.ch}

\icmlkeywords{Machine Learning, Provable Robustness, Robustness Certification, Adversarial Attack, Resnet18, Adversarial Example, ICML}

\vskip 0.3in
]

\printAffiliationsAndNotice{} 

\begin{abstract}
We present a training system, which can provably defend significantly larger neural networks than previously possible, including ResNet-34 and DenseNet-100. Our approach is based on differentiable abstract interpretation and introduces two novel concepts: (i) abstract layers for fine-tuning the precision and scalability of the abstraction, (ii) a flexible domain specific language (DSL) for describing training objectives that combine abstract and concrete losses with arbitrary specifications.  Our training method is implemented in the DiffAI system.
\end{abstract}

\section{Introduction}

\begin{figure*}[!t]
\centering
\includegraphics[width=\textwidth]{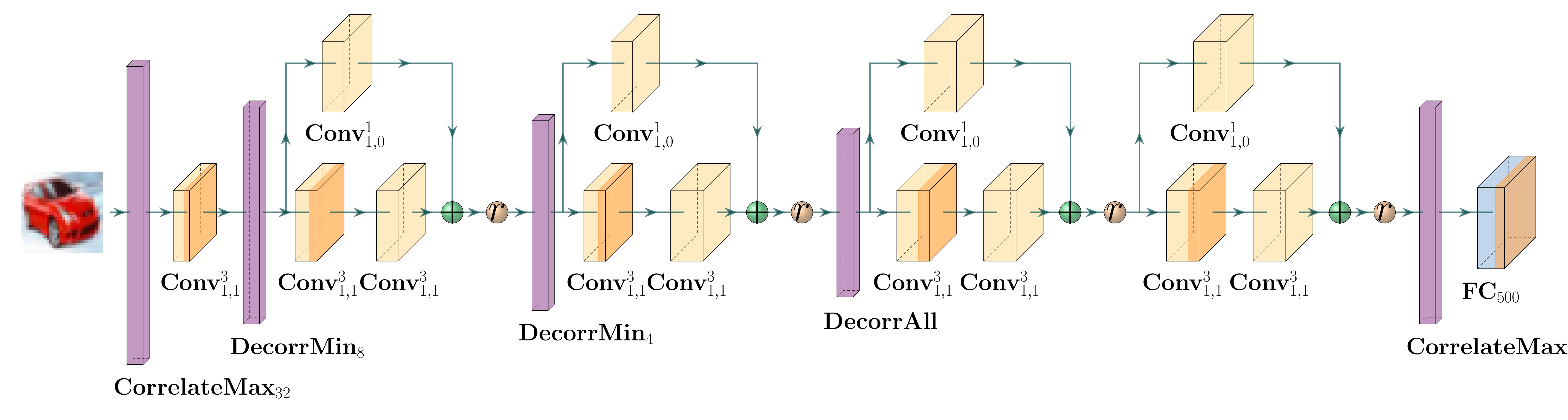}
\vspace{-1cm}
\caption{ResNet-Tiny with Abstract Layers. Layers with dark orange on their right include a ReLU, and the sphere with an $r$ in it is also a ReLU. Conv$^k_{s,p}$ is a convolution with a kernel size of $k$, a stride of $s$, and a padding $p$, This net has 311k neurons and 18m parameters.}
\label{Fi:resnetabslayer}
\end{figure*}

Recent work has shown that neural networks are susceptible to adversarial attacks \citet{adversarialDiscovery}: small, imperceptible perturbations which cause the network to misclassify the input. This has led to growing interest in training procedures to produce robust networks \cite{gu2014towards, zheng2016improving}, new adversarial attacks \cite{papernot2016limitations, moosavi2017universal, xiao2018generating, athalye2017synthesizing, evtimov2017robust}, as well as defenses which use these attacks during training \cite{goodfellow2014explaining, tramer2017ensemble, yuan2017adversarial, huang2015learning, madry2017towards, dong2018boosting}. While networks defended using attacks may be experimentally robust, it has been shown that in general more data is needed \cite{schmidt2018adversarially} and that this style of training is sample inefficient \cite{khoury2018geometry}.
Further, while detecting advarsarial attacks \cite{rozsa2016adversarial, bhagoji2017dimensionality, feinman2017detecting, grosse2017statistical} appears a promising contingency, \citet{CarliniW17} found that many of these techniques were insufficient.

The list of possible attacks is extensive (e.g., \cite{akhtar2018threat}) and constantly expanding, motivating the need for methods which can ensure that neural networks are \emph{provably} robust against these attacks. \citet{katz2017reluplex} developed a neural network verification system based on SMT solvers, however it only scaled to small networks.  \citet{ai2} introduced abstract interpretation \cite{CC77} as a method for verifying much larger networks. However, as the size of networks that verification systems could handle increased, it became clear that verifiable robustness could be significantly improved by employing provably robust training. The first attempts for training provably robust networks \cite{raghunathan2018certified, kolter2017provable, dvijotham2018training} scaled to small sizes with at most two convolutional layers.  Later work saw the development of two methods: (i) the dual-method in the case of \citet{wong2018scaling}, and (ii) differentiable abstract interpretation introduced by \citet{diffai} (DiffAI) and used in \citet{ibp} (IBP) and \citet{mixtrain} (MixTrain). While these pushed the boundary in terms of provable verified robustness and network size (with networks of up to $230$k neurons), scaling a provable defense to a full ImageNet sized network remains a key challenge. In particular, ResNet-34 represents an important milestone to achieving this goal as it is the smallest residual network proposed by \citet{resnet18}.

To address this challenge, we introduce a novel approach to robustness, one where the network itself is designed to be provably robust similar to attempts which aim to design networks to be experimentally robust by construction \cite{cisse2017parseval, sabour2017dynamic}. In particular, we introduce the paradigm of ``programming to prove'', long known to the programming languages community \cite{lf, fastver}, as a technique for creating provably robust architectures. We show how to integrate this idea with DiffAI, resulting in a system than can train a provably robust ResNet-34 (a smaller resnet is shown in \figref{resnetabslayer}).

\textbf{Main Contributions} Our main contributions are:
\begin{itemize}
\item {The concept of an abstract layer which has no effect on standard network execution but improves provably robust learning.}
\item {A domain specific language (DSL) for specifying sophisticated training objectives.}
\item {A complete implementation and evaluation of our method. Our experimental results indicate the approach can achieve provable robustness for networks an order of magnitude larger than prior work.}
\end{itemize}

\section{Background on Robust Training}\label{Se:background}
We now provide necessary background on training neural networks to be provably robust against adversarial examples. A neural network $N_\theta\colon \mathbb{R}^d \rightarrow \mathbb{R}^k$ maps a $d$-dimensional input to a $k$-dimensional output based on learned weights $\theta$. Let $B_\epsilon(x)$ be the $\ell_\infty$-ball of radius $\epsilon$ around an input $x\in\mathbb{R}^d$. A network $N_\theta$ is called \emph{$\epsilon$-robust} around a point $x \in \mathbb{R}^d$ if $\forall \tilde x \in B_\epsilon(x), N_\theta(\tilde x)_i > N_\theta(\tilde x)_j$ where $i,j\in\{1,\ldots,k\}$ and $j \ne i$. The goal of a robust training procedure is to learn a $\theta$ such that: (i) $N_{\theta}$ assigns the correct class $y_i$ to each training example $x_i$, and (ii) $N_\theta$ is $\epsilon$-robust around each example $x_i$.

\paragraph{Differentiable Abstract Interpretation}
In this work we leverage the differentiable abstract interpretation framework introduced by \citet{diffai}. Here, one verifies neural network robustness and formulates provability losses by constructing sound overapproximations using \emph{abstract interpretation} \cite{CC77}. We now introduce the necessary terms used later in the paper.
\begin{definition}
An \emph{abstract domain} $\mathcal D$ consists of: (a) abstract elements representing a set of concrete points in $\mathcal{P}(\mathbb{R}^p)$ for $p\in\mathbb N$, (b) a \emph{concretization function} $\gamma\colon\mathcal D\to\mathcal{P}(\mathbb{R}^p)$ mapping an abstract element $d \in \mathcal D $ to the set of concrete points it represents, and (c) a set of abstract transformers $T^{\#}$ approximating the concrete transformer $T$ in $\mathcal{P}(\mathbb{R}^p)$, i.e., $T(\gamma(d)) \subseteq \gamma(T^\#(d))$.
\end{definition}

Our approach additionally requires the existence of an abstraction function $\alpha\colon\mathcal{P}(\mathbb{R}^p)\to\mathcal D$ mapping the set of concrete points in $\mathbb{R}^p$ to an abstract element $d \in \mathcal{D}$.
Abstract transformers compose and hence by defining abstract transformers for each basic operation in a neural network $N$, we can derive an overall abstract transformer $T^\#_N$ for the entire $N$. 
We apply abstract interpretation to compute $T^\#_N(\alpha(B_\epsilon(x)))$, describing a superset of the possible outputs of $N$ for all inputs in $B_\epsilon(x)$ which can be used to compute an abstract loss as in \citet{diffai}.

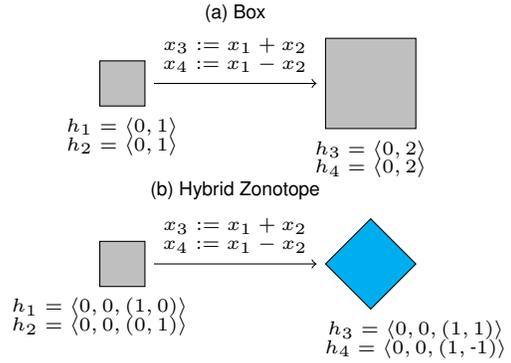
\begin{figure}

\centering
\begin{tikzpicture}[font=\tiny\sffamily, scale=1.2,every node/.style={scale=1.2}]

\pgfmathsetmacro{\cubex}{0.5}
\pgfmathsetmacro{\cubey}{0.5}
\pgfmathsetmacro{\cubez}{0.5}

\draw[fill=lightgray] (0,-0.25,0) -- ++(-\cubex,0,0) -- ++(0,-\cubey,0) -- ++(\cubex,0,0)--cycle;
\draw[fill=lightgray] (0,-2.25,0) -- ++(-\cubex,0,0) -- ++(0,-\cubey,0) -- ++(\cubex,0,0)--cycle;
\draw[fill=lightgray] (3,0,0) -- ++(-2*\cubex,0,0) -- ++(0,-2*\cubey,0) -- ++(2*\cubex,0,0)--cycle;
\draw[fill=cyan] (2,-2.5,0) -- ++(\cubex,\cubey,0) -- ++(\cubex,-\cubey,0) -- ++(-\cubex,-\cubey,0)--cycle;

\path [->] (0.1,-0.5) edge node[left,above, darkblue, line width=3mm] {} (1.9,-0.5);
\path [->] (0.1,-2.5) edge node[left,above, darkblue, line width=3mm] {} (1.9,-2.5);

\coordinate (x11) at (1,-0.5);
\node at (x11) [above=2.5mm  of x11] {$x_{3}:=x_{1}+x_{2}$};
\node at (x11) [above=0mm  of x11] {$x_{4}:=x_{1}-x_{2}$};

\coordinate (x11) at (1,-2.5);
\node at (x11) [above=2.5mm  of x11] {$x_{3}:=x_{1}+x_{2}$};
\node at (x11) [above=0mm  of x11] {$x_{4}:=x_{1}-x_{2}$};

\coordinate (x11) at (-0.25,-0.75);
\node at (x11) [below=0mm  of x11] {$h_{1}=\langle 0, 1 \rangle$};
\node at (x11) [below=2.5mm  of x11] {$h_{2}=\langle 0, 1\rangle$};

\coordinate (x11) at (2.5,-1);
\node at (x11) [below=0mm  of x11] {$h_{3}=\langle 0, 2 \rangle$};
\node at (x11) [below=2.5mm  of x11] {$h_{4}=\langle 0, 2\rangle$};

\coordinate (x11) at (-0.5,-2.75);
\node at (x11) [below=0mm  of x11] {$h_{1}=\langle 0, 0,(1,0) \rangle$};
\node at (x11) [below=2.5mm  of x11] {$h_{2}=\langle 0, 0,(0,1) \rangle$};

\coordinate (x11) at (3,-3);
\node at (x11) [below=0mm  of x11] {$h_{3}=\langle 0, 0,(1,1) \rangle$};
\node at (x11) [below=2.5mm  of x11] {$h_{4}=\langle 0, 0,(1,\text{-}1) \rangle$};

\coordinate (x11) at (1,0.5);
\node at (x11) [below=0mm  of x11] {(a) Box};

\coordinate (x11) at (1,-1.25);
\node at (x11) [below=2.5mm  of x11] {(b) Hybrid Zonotope};

\end{tikzpicture}
\caption{Comparing the precision of the affine transformers in the (a) Box and (b) Hybrid Zonotope domains. }
\label{Fi:transformers}
\end{figure}

\paragraph{Hybrid Zonotope Domain}
In this work we use the \emph{Hybrid Zonotope Domain} as described by \citet{diffai}.  This domain, introduced originally by \citet{perturbed}, is a generalization of two domains: (i) the simple Box domain (the Box domain is also referred to as interval bound propagation in \citet{ibp}) and, (ii) the base zonotope domain \citet{t1p}. The main benefit of hybrid zonotopes is that they allow for more fine-grained control of analysis precision and performance.

The Hybrid Zonotope domain associates with every computed result $v$ (e.g., a neuron) in the network, a triplet $h_v = \langle (h_C)_v, (h_B)_v, (h_E)_v \rangle$ where $h = \langle h_C, h_V, h_E \rangle$ and is referred to as the hybrid zonotope over all $p$ variables. Here, $(h_C)_v\in\mathbb{R}$ is a center point, $(h_B)_v\in\mathbb{R}_{\ge0}$ is a non-negative \emph{uncorrelated error coefficient} (similar to the Box domain), and  $(h_E)_v~\in~\mathbb{R}^m$ are the  \emph{correlated error coefficients} the number $m$ of which determine the accuracy of the domain.
These coefficients define an affine function $\widehat{h}$ which is parameterized by the \emph{correlated error terms} $e~\in~[-1,1]^m$ and an \emph{uncorrelated error term} $\beta~\in~[-1,1]^p$:
\[
\widehat{h}(\beta,e) = ({h_1}(\beta,e), \ldots, {h_p}(\beta,e))
\]
where:
\[
\widehat{h_v}(\beta,e) = (h_C)_v + (h_B)_v \cdot \beta_v + (h_E)_v \cdot e
\]
Different variables share the correlated error terms which introduces dependencies between variables making over-approximations more precise than those produced with the Box domain (which does not track dependencies).
Formally, the \emph{concretization} function $\gamma_\textit{H}$ of a hybrid zonotope $h$ is
\[
\gamma_\textit{H}(h) = \{ \widehat{h}(\beta,e) \mid \beta \in [-1,1]^p, e \in [-1,1]^m \}.
\]
A box $b$ can be expressed as a hybrid zonotope $h$ with $h_\textit{C}=b_\textit{C}$ (the box's center), $h_\textit{B}=b_\textit{B}$ (the box's radius) and $m=0$.
Descriptions of our hybrid zonotope transformers (e.g., ReLU), can be found in \citet{diffai}.

\paragraph{Interval concretization}
For operations such as constructing an abstract loss or building heuristics in abstract layers, it is necessary to determine the bounds of a hybrid zonotope $h$ for the $i$-th variable using \emph{interval concretization}:
\[
\iota_\textit{H}(h)_i = [ (h_\textit{C})_i - \epsilon_{\textit{H}}(h)_i ,\; (h_\textit{C})_i + \epsilon_{\textit{H}}(h)_i ]
\]
where $\epsilon_{\textit{H}}(h)_i = (h_\textit{B})_i+\sum_{j = 1}^{m} \left\lvert (h_\textit{E})_{i,j}\right\rvert$ is the \emph{total error}.

\paragraph{Example: Box vs. Hybrid Zonotope}
\figref{transformers} shows an affine transformation on inputs abstracted in both the Box and the Hybrid Zonotope domains. The box representation in \figref{transformers} (a) only contains the center and the uncorrelated error coefficients whereas the hybrid zonotope representation in \figref{transformers} (b) also contains non-zero correlated error coefficients. The affine transformation creates dependency between $x_{3}$ and $x_{4}$ as they are assigned values using affine expressions defined over the same variables $x_{1}$ and $x_{2}$. The Box domain cannot capture this and as a result its output is less precise (contains more concrete points) than the one produced with Hybrid Zonotope domain.

\section{Abstract Layers for Verifiable Networks}

Program verification often involves both the addition of erasable annotations
\cite{Sascha:08} and program transformations to make the resulting
(semantically equivalent program) more suitable for verification \cite{fastver}. That is, unlike
standard transformations which aim to produce a program that runs faster, here, the goal is to produce
a more verifiable program. Our key insight is to leverage this ``programming to prove''
paradigm in a similar fashion when designing robust neural networks.

Based on this insight, we describe the novel concept of \emph{Abstract Layers}. These layers
are specifically provided by the network designer but differ from traditional concrete layers in
that they have no effect on the concrete execution. Instead, they only affect the analysis of
the network, i.e., they only modify abstract elements that propagate through
the layers (e.g., boxes or hybrid zonotopes).

We describe two types of abstract layers designed to tune the precision and scalability
of the analysis with the Hybrid Zonotope domain. For all abstract layers, we
describe their effect on a given hybrid zonotope $h$ with $m$ correlated
error coefficients, producing a new hybrid zonotope $h'$. For our abstract  layers, it holds that $h'_C = h_C$.

\subsection{Correlation layers}
A correlation layer increases the precision of the analysis in successive layers by producing
a new hybrid zonotope $h'$ which contains more correlated error coefficients than the original input $h$.
We note that here we have $\gamma(h') = \gamma(h)$, that is, both hybrid zonotopes actually represent the same set of points.
However, the advantage of $h'$ over $h$ is that $h'$ contains more shared dependencies between different dimensions (variables) than $h$,
meaning that successive steps of the analysis using $h'$ will be produce more precise results than those same steps using $h$.

Informally, a correlation layer selects a set of dimension indices (variables) $P$ and creates $|P|$ new correlated
error coefficients. For each selected variable $i \in P$, we introduce one correlated error coefficient whose value is that of
the variable's uncorrelated error coefficient. All other remaining correlated coefficients (a total of $|P| - 1$) for $i$ are set to 0.
For all variables not in $P$, their new correlated error coefficients are all set to $0$. More formally, given $h$, we define $h'$ as follows:

\begin{tabular}{lccr}
 $h'_{B,i} = h_{B,i}$ & & & $i \notin P$\\
 $h'_{B,i} = 0$ & & & $i \in P$ \\
 $h'_{E,i,j} = h_{E,i,j}$   & &  & $\forall 0 \le i < p, 0 \le j < m$\\
 $h'_{E,i,m + t} = h_{B,i}$ & & & $i \in P \wedge t = |P_{< i}|$ \\
 $h'_{E,i,m + t} = 0$       & & & $\forall t < |P| . i \notin P \vee t \neq |P_{< i}|$ \\
\end{tabular}

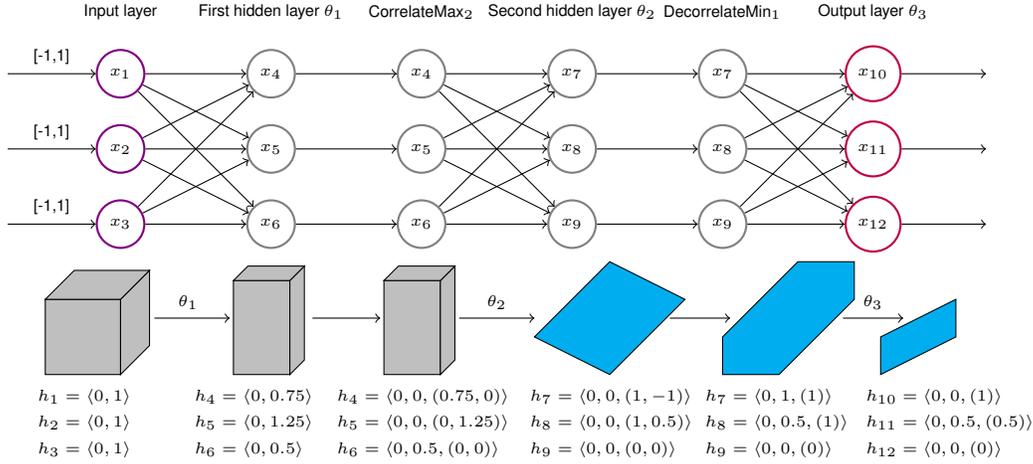
\begin{figure*}

\centering
\begin{tikzpicture}[font=\tiny\sffamily, scale=1,every node/.style={scale=1}]
    \node[shape=circle,draw=violet, thick] (A) at (0,0) {$x_{1}$};
    \node[shape=circle,draw=violet, thick] (B) at (0,-1) {$x_{2}$};
    \node[shape=circle,draw=violet, thick] (C) at (0,-2.0) {$x_{3}$};

    \node[shape=circle,draw=gray, thick] (D) at (2,0) {$x_{4}$};
    \node[shape=circle,draw=gray, thick] (E) at (2,-1) {$x_{5}$};
    \node[shape=circle,draw=gray, thick] (F) at (2,-2) {$x_{6}$};

    \node[shape=circle,draw=gray, thick] (G) at (4,0) {$x_{4}$};
    \node[shape=circle,draw=gray, thick] (H) at (4,-1) {$x_{5}$};
    \node[shape=circle,draw=gray, thick] (I) at (4,-2) {$x_{6}$};

    \node[shape=circle,draw=gray, thick] (J) at (6,0) {$x_{7}$};
    \node[shape=circle,draw=gray, thick] (K) at (6,-1) {$x_{8}$};
    \node[shape=circle,draw=gray, thick] (L) at (6,-2) {$x_{9}$};

    \node[shape=circle,draw=gray, thick] (M) at (8,0) {$x_{7}$};
    \node[shape=circle,draw=gray, thick] (N) at (8,-1) {$x_{8}$};
    \node[shape=circle,draw=gray, thick] (O) at (8,-2) {$x_{9}$};

    \node[shape=circle,draw=purple, thick] (P) at (10,0) {$x_{10}$} ;
    \node[shape=circle,draw=purple, thick] (Q) at (10,-1) {$x_{11}$} ;
    \node[shape=circle,draw=purple, thick] (R) at (10,-2.0) {$x_{12}$} ;

    \path [->] (A) edge node[left,above, thick] {} (D);
    \path [->](B) edge node[left,below, thick] {} (E);
    \path [->](C) edge node[left,below, thick] {} (F);

    \path [->](A) edge node[pos=0.25,left,above, thick] {} (E);
    \path [->](B) edge node[pos=0.25,left,above, thick] {} (F);
    \path [->](C) edge node[pos=0.25,left,above, thick] {} (D);

    \path [->](A) edge node[pos=0.25,left,above, thick] {} (F);
    \path [->](B) edge node[pos=0.25,left,above, thick] {} (D);
    \path [->](C) edge node[pos=0.25,left,above, thick] {} (E);

    \path [->] (D) edge node[left,above, thick] {} (G);
    \path [->](E) edge node[left,below, thick] {} (H);
    \path [->](F) edge node[left,below, thick] {} (I);

    \path [->] (G) edge node[left,above, thick] {} (J);
    \path [->](H) edge node[left,below, thick] {} (K);
    \path [->](I) edge node[left,below, thick] {} (L);

    \path [->](G) edge node[pos=0.25,left,above, thick] {} (K);
    \path [->](H) edge node[pos=0.25,left,above, thick] {} (L);
    \path [->](I) edge node[pos=0.25,left,above, thick] {} (J);

    \path [->](G) edge node[pos=0.25,left,above, thick] {} (L);
    \path [->](H) edge node[pos=0.25,left,above, thick] {} (J);
    \path [->](I) edge node[pos=0.25,left,above, thick] {} (K);

    \path [->] (J) edge node[left,above, thick] {} (M);
    \path [->](K) edge node[left,below, thick] {} (N);
    \path [->](L) edge node[left,below, thick] {} (O);

  \path [->] (M) edge node[left,above, thick] {} (P);
    \path [->](N) edge node[left,below, thick] {} (Q);
    \path [->](O) edge node[left,below, thick] {} (R);

    \path [->](M) edge node[pos=0.25,left,above, thick] {} (Q);
    \path [->](N) edge node[pos=0.25,left,above, thick] {} (R);
    \path [->](O) edge node[pos=0.25,left,above, thick] {} (P);

    \path [->](M) edge node[pos=0.25,left,above, thick] {} (R);
    \path [->](N) edge node[pos=0.25,left,above, thick] {} (P);
    \path [->](O) edge node[pos=0.25,left,above, thick] {} (Q);

\path [->] (-1.5,0) edge node[left,above, thick] {[-1,1]} (A);
\path [->](-1.5,-1) edge node[left,above, thick] {[-1,1]} (B);
\path [->](-1.5,-2) edge node[left,above, thick] {[-1,1]} (C);

\path [->] (P) edge node[left,above, thick] {} (11.5,0);
\path [->](Q) edge node[left,below, thick] {} (11.5,-1);
\path [->](R) edge node[left,below, thick] {} (11.5,-2.0);

\coordinate (H) at (A);
\node at (H) [above=6mm  of H] { Input layer };

\coordinate (H) at (D);
\node at (H) [above=6mm  of H] { First hidden layer $\theta_{1}$ };

\coordinate (H) at (G);
\node at (H) [above=6mm  of H] { CorrelateMax$_{2}$ };

\coordinate (H) at (J);
\node at (H) [above=6mm  of H] { Second hidden layer $\theta_{2}$ };

\coordinate (H) at (M);
\node at (H) [above=6mm  of H] { DecorrelateMin$_{1}$ };

\coordinate (H) at (P);
\node at (H) [above=6mm  of H] { Output layer $\theta_{3}$};

\pgfmathsetmacro{\cubex}{1}
\pgfmathsetmacro{\cubey}{1}
\pgfmathsetmacro{\cubez}{1}
\definecolor{lightblue}{rgb}{0.0, 0.18, 0.65}
\draw[fill=lightgray] (0,-3,0) -- ++(-\cubex,0,0) -- ++(0,-\cubey,0) -- ++(\cubex,0,0)--cycle;
\draw[fill=lightgray] (0,-3,0) -- ++(0,0,-\cubez) -- ++(0,-\cubey,0) -- ++(0,0,\cubez) -- cycle;
\draw[fill=lightgray] (0,-3,0) -- ++(-\cubex,0,0) -- ++(0,0,-\cubez) -- ++(\cubex,0,0) -- cycle;

 \coordinate (x1) at (-0.5,-4,0);
\node at (x1) [below=0.5mm  of x1] { $\begin{aligned} & h_{1}=\langle 0, 1\rangle \\
							&h_{2}=\langle 0,1 \rangle\\
							&h_{3}=\langle 0,1 \rangle \end{aligned}$ };

\coordinate (x1) at (0.9,-3.25);
\node at (x1) [above=0mm  of x1] {$\theta_{1}$};

\pgfmathsetmacro{\cubex}{0.75}
\pgfmathsetmacro{\cubey}{1.25}
\pgfmathsetmacro{\cubez}{0.5}

\draw[fill=lightgray] (2.25,-2.75,0) -- ++(-\cubex,0,0) -- ++(0,-\cubey,0) -- ++(\cubex,0,0)--cycle;
\draw[fill=lightgray] (2.25,-2.75,0) -- ++(0,0,-\cubez) -- ++(0,-\cubey,0) -- ++(0,0,\cubez) -- cycle;
\draw[fill=lightgray] (2.25,-2.75,0) -- ++(-\cubex,0,0) -- ++(0,0,-\cubez) -- ++(\cubex,0,0) -- cycle;

\path [->] (0.45,-3.25) edge node[left,above, darkblue, line width=3mm] {} (1.45,-3.25);

\coordinate (x1) at (1.75,-4,0);
\node at (x1) [below=0.5mm  of x1] { $\begin{aligned} & h_{4}=\langle 0, 0.75\rangle \\
							&h_{5}=\langle 0,1.25 \rangle\\
							&h_{6}=\langle 0,0.5 \rangle \end{aligned}$ };

\draw[fill=lightgray] (4.25,-2.75,0) -- ++(-\cubex,0,0) -- ++(0,-\cubey,0) -- ++(\cubex,0,0)--cycle;
\draw[fill=lightgray] (4.25,-2.75,0) -- ++(0,0,-\cubez) -- ++(0,-\cubey,0) -- ++(0,0,\cubez) -- cycle;
\draw[fill=lightgray] (4.25,-2.75,0) -- ++(-\cubex,0,0) -- ++(0,0,-\cubez) -- ++(\cubex,0,0) -- cycle;

\path [->] (2.55,-3.25) edge node[left,above, darkblue, line width=3mm] {} (3.45,-3.25);

\coordinate (x1) at (4,-4,0);
\node at (x1) [below=0.5mm  of x1] { $\begin{aligned} & h_{4}=\langle 0, 0, (0.75, 0)\rangle \\
							&h_{5}=\langle 0,0, (0,1.25) \rangle\\
							&h_{6}=\langle 0,0.5, (0,0) \rangle \end{aligned}$ };

\pgfmathsetmacro{\cubex}{1}
\pgfmathsetmacro{\cubey}{1}
\pgfmathsetmacro{\cubez}{1}
\draw[fill=cyan] (5.5,-3.5,0) -- ++(\cubex,\cubey,0) -- ++(\cubex,-\cubey/2,0) -- ++(-\cubex,-\cubey,0)--cycle;

\path [->] (4.5,-3.25) edge node[left,above, darkblue, line width=3mm] {} (5.5,-3.25);

\coordinate (x1) at (5,-3.25);
\node at (x1) [above=0mm  of x1] {$\theta_{2}$};

\coordinate (x1) at (6.5,-4);
\node at (x1) [below=0.5mm  of x1] { $\begin{aligned} & h_{7}=\langle 0, 0, (1, -1)\rangle \\
							&h_{8}=\langle 0,0, (1,0.5) \rangle\\
							&h_{9}=\langle 0,0, (0,0) \rangle \end{aligned}$ };

\draw[fill=cyan] (8,-3.5,0) -- ++(\cubex,\cubey,0) -- ++(0.75*\cubex,0,0) -- ++(0,-0.5*\cubey,0) -- ++(-\cubex,-\cubey,0) -- ++(-0.75*\cubex,0,0) -- ++(0,0.5*\cubey,0)--cycle;
\path [->] (7.3,-3.25) edge node[left,above, darkblue, line width=3mm] {} (8.1,-3.25);

\coordinate (x1) at (8.7,-4);
\node at (x1) [below=0.5mm  of x1] { $\begin{aligned} & h_{7}=\langle 0, 1, (1)\rangle \\
							&h_{8}=\langle 0,0.5, (1) \rangle\\
							&h_{9}=\langle 0,0, (0) \rangle \end{aligned}$ };

\pgfmathsetmacro{\cubex}{1}
\pgfmathsetmacro{\cubey}{1}
\pgfmathsetmacro{\cubez}{1}
\draw[fill=cyan] (10.1,-3.5,0) -- ++(\cubex,\cubey/2,0) -- ++(0,-\cubey/2,0) -- ++(-\cubex,-\cubey/2,0)--cycle;
\path [->] (9.6,-3.25) edge node[left,above, darkblue, line width=3mm] {} (10.4,-3.25);
\coordinate (x1) at (10,-3.25);
\node at (x1) [above=0mm  of x1] {$\theta_{3}$};

\coordinate (x1) at (11,-4);
\node at (x1) [below=0.5mm  of x1] { $\begin{aligned} & h_{10}=\langle 0, 0, (1)\rangle \\
							&h_{11}=\langle 0,0.5, (0.5) \rangle\\
							&h_{12}=\langle 0,0, (0) \rangle \end{aligned}$ };


\end{tikzpicture}
\caption{Our analysis on an example toy network augmented with abstract layers.}
\label{Fi:nn2}
\end{figure*}

Here we use $P_{< i}$ to denote the subset of $P$ where each element is smaller than $i$.
Next, we define four variants of a correlation layer based on the choice of the set $P$.

\textbf{CorrelateAll} correlates all uncorrelated coefficients in all $p$ dimensions thereby adding $p$ correlated error coefficients. Formally, it uses $P = \{ i~~|~~0 \leq i < p \}$.

\textbf{CorrelateFixed$_k$} correlates $k$ fixed dimensions, chosen by taking every $\frac{p}{k}$ of the flattened list of dimension indices. Formally, we have $P = \{ \lfloor\frac{i \cdot p}{k}\rfloor~~|~~0 \leq i < k\}$.

\textbf{CorrelateMax$_k$} correlates the first $k$ dimensions whose interval concretization (see \secref{background}) has the largest upper bound value. This heuristic aims to improve precision while still keeping the analysis scalable.
Formally, we have $P = \{ i~~|~~\textsc{UB}(\iota_\textit{H}(h)_i) \in \textsc{top}_{k}(\textsc{UB}(\iota_\textit{H}(h))) \}$ where $\textsc{top}_{k}$ returns the $k$ largest elements and \textsc{UB} returns the upper bound of an interval.

\textbf{CorrelateMaxPool$_{c,w,h,s}$} correlates dimensions chosen using MaxPooling \cite{krizhevsky2012imagenet}. We apply MaxPooling with kernel size $(c,w,h)$ and stride $s$ on a function $f$ defined over $h$. If the
correlation is applied before the first layer then $f=h_C$ otherwise $f=h_B$. Formally, $P = \{i~~|~~ f(i) \in \textsc{Maxpool}_{c,w,h,s}(f) \}$.

\subsection{Decorrelation layers}

The purpose of decorrelation layers is opposite that of correlation layers: to reduce the number of correlated coefficients so to make analysis for successive layers more efficient but less precise. Concretely, a decorrelation layer removes correlated error coefficients and adds their absolute sum to the value of uncorrelated error coefficients in each dimension. We now introduce two choices for the set $P$, each defining different dimensions to be decorelated:

\textbf{DecorrelateAll} produces a hybrid zonotope with no correlated coefficients in any dimension and in each dimension, the uncorrelated coefficient is defined as: 
$$h'_{B,i} = h_{B,i} + \sum_{j=0}^{m-1} |h_{E,i,j}|$$
\textbf{DecorrelateMin$_k$} is based on a heuristic to minimize the loss of
precision due to decorrelation by removing $m - k$ correlated error coefficients whose absolute sum in all dimensions is the smallest. As a result,  $k$ correlated coefficients remain in the output. Formally, we define $P$ to be the indices of the $m - k$ smallest elements of the sequence $\{ \sum_{i=0}^{p-1} |h_{E,i,j}| \}_{j=0\ldots m-1}$.  Then, the new zonotope $h'$, where $i \in [0,p)$, is defined as:

\begin{tabular}{lrrr}
$h'_{B,i} = h_{B,i} + \sum_{j \in P} |h_{E,i,j}|$ & & & \\
\\
$h'_{E,i,j} = h_{E,i,t}$ & $t \notin P \wedge j = t - |P_{< t}|$&  &
\end{tabular}

Informally, in the first equation, we accumulate all correlated coefficients chosen for removal (that is, the set $P$) into the uncorrelated coefficient, while the second equation ensures the remaining correlated coefficients are shifted to be next to each other (order is preserved).

\subsection{DeepLoss}

For deeper neural networks such as ResNet-18, it is possible for naive abstraction imprecision to grow exponentially to the point where overflow occurs before the final loss is calculated, making optimizing that loss futile.
In such cases, we would like the network to produce more precise results in intermediate layers, before an overflow occurs.
As these layers do not have the same number of neurons as target classifications, we cannot optimize using a standard
provability loss.  Instead, a loss on a generic heuristic for provability must be defined on the output of a specific layer.
As this loss does not effect concrete execution and operates using the abstract element from a specific layer, we also consider it a form of an abstract layer.
We define the following losses on an interval concretization $c$ in $n$ dimensions:

\begin{tabular}{lcr}
$L_{\text{lb},f,i}(c)$ & = & $\max \{f(c_{j,2} - c_{i,1})~~|~~c_{j,1} \leq c_{i,1} \}$\\
$L_{\text{ub},f,i}(c)$ & = & $\max \{f(c_{i,2} - c_{j,1})~~|~~c_{i,2} \leq c_{j,2} \}$\\
$L_{\text{deep},f}(d)$ & = & $\frac{1}{2n} \sum_{i=0}^{n-1} (L_{\text{lb},f,i}(\iota(d)) + L_{\text{ub},f,i}(\iota(d)))$
\end{tabular}

where $f$ is a positive activation and $L_{\text{deep},f}$ combines the first two losses for each dimension of an arbitrary abstract element $d$. Intuitively, this loss measures and sums for each dimension the worst offending overlap between the concretization lower bound in that dimension and the upper bound of any other dimension, and visa versa.

In our experiments, we used ReLU for $f$. A naive implementation of the above loss would require $n^2$ computations (and potentially $n^2$ space on a GPU), which could be problematic given that the loss is intended to be used on the output of an intermediate, and presumably quite wide, layer. While a matrix multiplication would also typically involve using up to an $n^2$ sized matrix, this loss is intended to be used between convolutions which typically permits significantly wider outputs through utilizing significantly smaller kernels. We implement it leveraging one dimensional MaxPool and Sort so that marshaling between the CPU and GPU is not required, and such that algorithmic optimizations to MaxPool can be leveraged to potentially\footnote{Provided an optimal implementation of MaxPool} provide an $O(n \log n)$ implementation.

\subsection{Example network with abstract layers}
\figref{nn2} shows our analysis with abstract layers on an example toy feedforward network. The neural network contains three neurons per layer. We add a CorrelateMax$_{2}$ and a DecorrelateMin$_{1}$ abstract layers after the first and second hidden layer respectively. We show the 3-dimensional shapes propagated through each layer along with the corresponding hybrid zonotope based encoding. The top, middle, and bottom neurons in each layer represent the x,y, and z-directions in the shapes. Our analysis abstracts the input region with a Box and propagates it through the first hidden layer.  After correlations are added, the abstraction is shown in blue (before correlations, the shape is gray).

\textbf{CorrelateMax$_{2}$} changes the encoding of the abstract element obtained after the first hidden layer by creating correlated error coefficients for neurons $x_{4}$ and $x_{5}$ whose upper bound is larger than that for $x_{6}$. The introduction of correlated error coefficients increases the precision of the result obtained \emph{after} applying the transformers for the second hidden layer as the resulting shape is no longer a box. We note that neuron $x_{9}$ is set to $0$ in the result.

\begin{table*}[t]\footnotesize
\centering
\caption{Our different goal constructors for training.  Each assumes the existence of a target label $t$. }
\vskip 0.1in
{ \footnotesize
\begin{tabular}{@{}rl l ll@{}}
\toprule
Goal constructor   & & abstract element $d=\alpha$($l,u$) & & loss($d,t$) \\
\midrule
Point  & & $\frac{l+u}{2}$ & & cross-entropy($d,t$) \\
Normal & & $\textsc{max}(\textsc{min}(0.5 \cdot (u - l) \cdot \textsc{normal\_rand}(0) + l, l), u)$ & & cross-entropy($d,t$)\\
Uniform  & & $\textsc{max}(\textsc{min}(0.5 \cdot (u - l) \cdot (\textsc{uniform\_rand}(2) - 1)  + l, l), u)$ & & cross-entropy($d,t$)\\
IFGSM$_k$ & & $\textsc{fgsm}$($k,l,u,t$) & & cross-entropy($d,t$) \\
Box        & & $\langle \frac{l+u}{2}, \frac{l-u}{2},0 \rangle$ & & cross-entropy($d,t$) \\
Mix($g_1$, $g_2$, $\lambda$) & & $(d_{1},d_{2})=\alpha^{g_{1}}$ ($l,u$) $\times \alpha^{g_{2}}$($l,u$) & & ($1$- $\lambda$)$\cdot \text{loss}^{g_{1}}$($d_{1},t$)+$\lambda \cdot \text{loss}^{g_{2}}$($d_{2},t$)\\
Sub($\delta$, $g$) & & $\alpha^{g}(0.5 \cdot (u + l - \delta \cdot (u - l)),0.5 \cdot (u + l + \delta \cdot (u - l),t)$ & & $\text{loss}^{g}$($d,t$)\\
Sample($\delta$, $r$, $g_{s}$, $g_{t}$)& &  $\alpha^{g_{t}}$($b - 0.5 \cdot \delta(u - l),b + 0.5 \cdot \delta(u - l),t$) where  & & $\text{loss}^{g_{t}}$($d,t$)\\
& & \quad $b=\textsc{center}(\text{Sub}$($1-r \cdot \delta,g_{s})$) & & \\
BiSample($g_{1}$,$g_{2}$) && $\alpha^{g_{2}}$($l',u',t$) where  && $\text{loss}^{g_{2}}$($d,t$)\\
& & \quad $l'=0.5 \cdot (l+u) - |\textsc{UB}$($\alpha^{g_{1}}(l,u))$- $0.5 \cdot (l+u)|$ and & & \\
& & \quad $u'=0.5 \cdot (l+u) + |\textsc{UB}$($\alpha^{g_{1}}(l,u))$- $0.5 \cdot (l+u)|$ & & \\
\bottomrule
\end{tabular}
}
\label{Ta:goal_constructors}
\end{table*}

\textbf{DecorrelateMin$_{1}$} removes the second correlated error coefficient as its absolute sum over all neurons is smaller. The absolute value of this coefficient is added to the uncorrelated coefficient in each dimension. This changes the concretization of the abstract element by removing dependencies making it less imprecise while increasing scalability. The output layer transformations next produce a result more precise than the box obtained without the abstract layers.

\begin{figure}

\centering
\begin{tikzpicture}[font=\tiny\sffamily, scale=1.4,every node/.style={scale=1.4}]

\newcommand\Square[1]{+(-#1,-#1) rectangle +(#1,#1)}

\draw[cyan, fill=cyan] (0,0) \Square{0.5};

\draw[black, fill=orange] (0.1,0.3) \Square{0.1};
\draw[black, fill=orange] (-0.2,0.37) \Square{0.1};
\draw[black, fill=orange] (-0.4,-0.1) \Square{0.1};
\draw[black, fill=orange] (-0.25,-0.24) \Square{0.1};
\draw[black, fill=orange] (0.35,-0.39) \Square{0.1};

\draw[cyan, fill=cyan] (3,0) \Square{0.5};

\draw[black, fill=orange] (3 + 0.1, 0.3) \Square{0.175};
\draw[black, fill=orange] (3 - 0.2, 0.25) \Square{0.175};
\draw[black, fill=orange] (3 - 0.25, -0.15) \Square{0.175};
\draw[black, fill=orange] (3 + 0.05, -0.33) \Square{0.175};
\draw[black, fill=orange] (3 + 0.25, -0.15) \Square{0.175};

\coordinate (x11) at (0,1);
\node at (x11) [below=0mm  of x11] {(a) $\delta=0.2$};

\coordinate (x11) at (3,1);
\node at (x11) [below=0mm  of x11] {(a) $\delta=0.35$};

\end{tikzpicture}
\vspace{-0.3cm}
\caption{Five two-dimensional abstractions (orange boxes) produced by $\alpha^g$ where $g$ is $\text{Sample}(\delta,\text{Normal},\text{Box})$ }
\vspace{-0.4cm}
\label{Fi:samples}
\end{figure}

\begin{table*}[t]
\centering
\caption{ Training schemes used for the experiments. Those with subscript 18 or L are used by ResNet-18 or ResNetLarge respectively.  }
\vskip 0.1in
{ \footnotesize

\begin{tabu}{@{}rl@{}}
\toprule
Name
& Training Scheme
\\ \midrule
Baseline & Mix(Point, Sub(Lin(0, 1, 150, 10), Box), Lin(0, 0.5, 150, 10)) \\
InSamp & Mix(Point, Sample(Lin(0, 1, 150, 10), 0.5, Normal, Box), Lin(0, 0.5, 150, 10)) \\
InSampLPA & Mix(Point, Sub(Lin(0,1,150,10), Sample(Lin(0,1,150,10), 0.5, Normal, Box)), Lin(0, 0.5, 150, 10)) \\
Adv$_k$IS & Mix(Sub(Lin(0, 1, 20, 20), IFGSM$_k$), Sample(Lin(0, 1, 150, 10), 0.5, Normal, Box), Lin(0, 0.5, 150, 10)) \\
Adv$_k$ISLPA & Mix(Sub(Lin(0, 1, 20, 20), IFGSM$_k$), \\
                 & \quad Sub(Lin(0,1,150,10),Sample(Lin(0, 1, 150, 10), 0.5, Normal, Box)), Lin(0, 0.5, 150, 10)) \\ 
Adv$_k$ISLPAUS & Mix(Sub(Lin(0, 1, 20, 20), IFGSM$_k$), \\
               & \quad Sub(Lin(0,1,150,10),Sample(Lin(0, 1, 150, 10), Uniform$_1$, Box)), Lin(0, 0.35, 150, 10)) \\ 

Baseline$_{S18}$ & Mix(Point, Sub(Lin(0, 1, 200, 40), Box), Lin(0, 0.5, 200, 40)) \\
InSamp$_{S18}$ & Mix(Point, Sample(Lin(0, 1, 200, 40), 0.5, Normal, Box), Lin(0, 0.5, 200, 40)) \\
Adv$_k$IS$_{S18}$ & Mix(Sub(Lin(0, 1, 20, 20), IFGSM$_k$), Sample(Lin(0, 1, 200, 40), 0.5, Normal, Box), Lin(0, 0.5, 200, 40)) \\ 

Baseline$_{S18}$ & Mix(Point, Sub(Lin(0, 1, 200, 40), Box), Lin(0, 0.5, 200, 40)) \\
InSamp$_{S18}$ & Mix(Point, Sample(Lin(0, 1, 200, 40), 0.5, Normal, Box), Lin(0, 0.5, 200, 40)) \\
Adv$_k$IS$_{S18}$ & Mix(Sub(Lin(0, 1, 20, 20), IFGSM$_k$), Sample(Lin(0, 1, 200, 40), 0.5, Normal, Box), Lin(0, 0.5, 200, 40)) \\ 

Adv$_k$ISLPA$_{R18}$ & Mix(Sub(Lin(0, 1, 20, 20), IFGSM$_k$), \\
                     & \quad Sub(Lin(0,1,200,40),Sample(Lin(0, 1, 200, 40), 1, Uniform, Box)), Lin(0, 0.5, 200, 40)) \\ 
InSampLPA$_{R34}$ & Mix(Point, Sub(Lin(0,1,200,40), Sample(Lin(0, 1, 200, 40), 1, Uniform, Box)), Lin(0, 0.5, 200, 40)) \\ 

Adv$_k$ISLPA$_{D100}$ & Mix(IFGSM$_k$, Sub(Lin(0,1,150,50),Sample(Lin(0, 1, 150, 50), 1, Uniform, Box)), Lin(0, 0.5, 150, 50)) \\ 

BiAdv$_L$ & Mix(IFGSM$_2$, BiSample(Sub(Lin(0, 1, 150, 30), IFGSM$_3$), Lin(0, 0.6, 200, 30)) \\
\bottomrule
\end{tabu}
}
\label{tab:methods}
\end{table*}

\section{Specifying Training Objectives}
We introduce a domain specific language (DSL) for specifying training objectives and parameter scheduling. For example, it can capture the training loss and scheduling proposed by \citet{ibp}~(IBP).

\subsection{Specifying Schedules}

We describe two constructors for describing a schedule used to adjust the values of training parameters (e.g., size of the balls around images used in training) dynamically, leading to improved results. A schedule is a function which uses the current training time step corresponding to the fractional number of epochs completed (e.g., completing 25000 of the 50000 examples from the first epoch on CIFAR10 would provide a time-step value of 0.5). The constructors below describe how to (recursively) build this function.

\textbf{Lin($a, b, m, n$)} specifies the parameter value should be the start value $a$ for the first $m$ epochs. Then, linear parameter annealing between start value $a$ and end value $b$ over $n$ epochs should be used to determine the parameter value.

\textbf{Until($m,s_{1},s_{2}$)} specifies that the first-schedule constructor $s_{1}$ will be used to determine the parameter value until $m$ epochs are reached, and then the second-schedule $s_{2}$ will be used but will be given the time with $m$ epochs subtracted.

\subsection{Specifying Training Goals}

We next describe the goal-constructors for describing how to build the abstraction function and training loss. At timestep $s$ of training a network $N$ on an example $o$ with a target label ($t$), for each goal constructor ($g$) in the abstract syntax tree (AST), we build: (i) an abstraction function ($\alpha^g$) which takes the input box for training specified by the lower ($l$) and upper bound ($u$) vectors as input and returns an abstract element $d=\alpha^g(l,u)$, and (ii) a loss function loss$^g$($d,t$). Before training, the user provides the goal which is parsed into an AST ($g_U$) and a training width ($\epsilon$). The loss used to train is, for a dataset with values in the range of $a$ to $b$:
\[
\text{loss}^{g_U}(T^\#_N(\alpha^{g_U}(\text{max}(o - \epsilon, a),\text{min}(o + \epsilon, b))), t).
\]
\tabref{goal_constructors} formalizes our goal constructors, described below:

\textbf{Point} returns the center of the input box specified by $l$ and $u$ for training and uses the cross entropy loss.

\textbf{Normal} returns a point sampled from the normal distribution around the input box (via the function \textsc{normal\_rand}) and clipped to that box. It uses the cross entropy loss.

\textbf{IFGSM$_k$} uses $k$ iterations of FGSM to find an adversarial example in the input box and uses the corresponding point for training. The cross entropy loss is used.

\textbf{Box} returns a hybrid zonotope abstract element with no correlated error coefficients abstracting the box between the lower ($l$) and upper-bound ($u$).
The loss concretizes the abstract element $d$ and returns the maximum cross entropy loss on a point in the concretization \cite{ibp}.

\textbf{Mix($g_1, g_2, \lambda$)} takes two goal constructors $g_1$ and $g_2$ and a float $\lambda$ as inputs. The abstract element used for training is the cartesian product of the abstractions $d_{1}$ and $d_{2}$ of the input box in $g_1$ and $g_{2}$. The loss linearly combines the loss functions from $g_{1}$ and $g_{2}$ using $\lambda$.

\textbf{Sub($\delta, g$)} takes a float $\delta$ and a goal constructor $g$ as inputs. It computes the abstract element for training by calling the abstraction function $\alpha^{g}$ of the constructor $g$ using the new bounds $l'= 0.5 \cdot (u + l - \delta \cdot (u - l))$ and $u'= 0.5 \cdot (u + l + \delta \cdot (u - l))$. The insight behind this constructor is to use training elements constructed from boxes that overlap with the input box. The output loss is the loss from $g$.

\textbf{Sample($\delta, g_{s}, g_{t}$)} uses Sub($1 - \delta$, $g_{s}$) to find a point $b$, by taking the center of the returned training element,
and passes $l' = b - 0.5 \cdot \delta(u - l)$  and $u' = b + 0.5 \cdot \delta(u - l)$ to the abstraction function $\alpha^{g_{t}}$ of $g_{t}$. The output loss is from $g_{t}$.  This is visualized in \figref{samples}.

\textbf{BiSample($g_{1},g_{2}$)} uses the abstract element $\alpha^{g_{1}}(l,u,t)$ for the input in $g_{1}$ and computes  ${l'= 0.5 \cdot (l+u) - |(\textsc{UB}(\alpha^{g_{1}}(l,u)))- 0.5 \cdot (l+u)|}$ and $~u'=0.5 \cdot (l+u) + |(\textsc{UB}(\alpha^{g_{1}}(l,u)))- 0.5 \cdot (l+u)|~$. The output element is $\alpha^{g_{2}}(l',u',t)$. It uses $g_{2}$'s loss.

We note that floating point parameters such as $\delta$ and $\lambda$ used in the constructors above can use scheduling constructors.

\subsection{Example Training Schemes}

Earlier, we observed that our training DSL could be used to specify complex training schemes such as IBP.  In particular, IBP uses linear parameter annealing on both the epsilon used in training and the weight of the provability loss, together with a cross entropy based loss function instead of the hinge loss designed by \citet{diffai}~(DiffAI). Using this customization, IBP improves on the results of DiffAI while still using the interval domain for training as done by DiffAI. In our DSL, this training scheme could be written as: Mix(Point, Sub(Lin(0,1,150,10), Box), Lin(0,0.5, 150,10)).

In Table~\ref{tab:methods}, we show a number of example training schemes captured as expressions in our DSL. We found the following schemes to be particularly useful (these are evaluated next):

\textbf{InSamp} interpolates between training on random points in the $L_\infty$ $\epsilon$-Ball and an abstract box surrounding the example. The idea is that it might be easier to train a network on a point to be $\epsilon$ robust if instead of it being only $\epsilon - \mu$ robust already, every point around in the $\epsilon$ box around it is also $\epsilon - \mu$ robust for small $\mu>0$.

\textbf{InSampLPA} is the same as InSamp, but also uses scheduling for the size of the sampling domain, by surrounding it with Sub.  The idea is that using the sampling domain is a kind of adversarial training, and it might be easier for the network to learn the standard dataset first.

\begin{table*}[t] 
\centering
\caption{ List of networks, and the training schemes that achieved the highest provable robustness. }
\vskip 0.1in
{ \footnotesize
\begin{tabu}{@{}rrrrrrrr@{}}
\toprule
Net
& Neurons
& Params
& Abstract Layers
& Training Scheme
& s/epoch
& Accuracy \%
& Verified Robustness \%
\\ \midrule
ResNet-Tiny  & 312k & 18m & ManyFixed   & Adv$_1$ISLPAUS    & 303 & 40.2 & 23.2   \\
SkipNet-18    & 558k & 15m & None       & Adv$_5$IS$_{S18}$ & 260 & 28.4 & 21.2 \\
ResNet-Large & 639k & 66m & LargeCombo & BiAdv$_{L}$   & 527 & 38.1 & 3.0 \\
ResNet-18 & 558k & 15m    & None & Adv$_5$ISLPA$_{R18}$ & 233 & 32.3 & 22.3 \\
ResNet-34 & 967k & 25m    & None & InSampLPA$_{R34}$ & 176 & 35.1 & 19.5 \\
DenseNet-100 & 4.5m & 748k & None & Adv$_5$ISLPA$_{D100}$ & 727 & 36.4 & 21.9 \\
\bottomrule
\end{tabu} }
\label{tab:simple_nets}
\end{table*}

\section{Experimental Setup}

Our system, and the code for reproducing experiments, is publicly available at \href{https://github.com/eth-sri/diffai}{https://github.com/eth-sri/diffai}. We implemented this system using PyTorch-0.4.1. We ran all experiments using GeForce RTX 2080 Ti GPUs. We do not use weight normalization reparameterization and clipping.

To demonstrate the effectiveness of our technique, we evaluate using the most challenging dataset commonly used for provable verification tasks, CIFAR-10 \cite{cifar}. We also use the largest commonly used epsilon, $\epsilon=0.031373 \sim 8/255$.  All accuracies and verifiable robustness percentages use the full 10,000 image test set. To augment the dataset and make it easier to learn, random cropping with a padding of 4 was used (this maintains image size) as well as random horizontal flipping.

While IBP and MixTrain presented improvements to robust training, we did not compare against these systems.  For IBP, the public code did not contain residual networks though we were able to integrate the proposed training improvements into DiffAI. For MixTain, the codebase was unavailable, and we chose not to perform normalization on the dataset prior to usage. Instead, we add a fixed layer to each network in order to make attack and verification epsilons easier to compare across different systems.

For testing the attacked accuracy, we used MI-FGSM \cite{dong2018boosting} with $\mu=0.8$, 20 iterations, and a step size of $0.0031373$.
To test verifiable robustness, we used DiffAI's built-in Hybrid-Zonotope domain (described earlier).

\subsection{ Evaluated Networks }

A brief overview of the network sizes we evaluate on and their training speed under well performing training schemes, is shown in Table~\ref{tab:simple_nets}. To the best of our knowledge, no other system can train as deep and as large provable networks as our system. In the Appendix, we provide the complete table for all training schemes. Next, we give a brief description of these networks and our training parameters.

\textbf{ResNet-Tiny} is a wide residual network, 12 layers deep, similar to the ResNet described by \citet{wong2018scaling} but with more and wider layers shown in \figref{resnetabslayer}.
It has 50\% more neurons than the largest CIFAR10 network trained via IBP or \citet{mixtrain}. For this network we always use an initial learning rate of $0.001$ with a schedule as used by IBP, and Adam optimization \cite{kingma2014adam}.  We also use an $L_2$ regularization constant of $0.01$.

\textbf{SkipNet-18} is an 18 layer deep network with 4 residual connections adapted from PyTorch's vision library.  
For this network we always use an initial learning rate of $0.1$ and a schedule where the rate is multiplied by $0.1$ at steps 10, 20, 250 and 300.
Instead of Adam, standard SGD is used. The $L_2$ regularization constant is set to $0.0005$.

\textbf{ResNet-18 and ResNet-34} are 18 and 34 layer (respectively) deep residual networks adapted from PyTorch's vision library.
For these network we always use an initial learning rate of $0.1$ and a similar schedule where the rate is multiplied by $0.1$ at steps 10, 20, 250, 300, and 350, and a batch size of 200.
We also use SGD here, but do not use any regularization.

\textbf{DenseNet-100} is a network with 99 layers, and many residual connections, adapted from the models proposed by \citet{huang2017densely}. 
To our knowledge, this is the largest network in terms of depth and the number of neurons to have been provably trained so far.
For this network, we always use an initial learning rate of $0.1$ and a schedule where the rate is multiplied by $0.1$ at steps 20, 50, 200, 250, and 300.
We also use SGD here, and no regularization, and a batch size of 50.  Due to its size, it is only verified using the Box domain and not with the hybrid Zonotope domain.

\subsection{ Abstract Layers }

To evaluate the effect of abstract layers, we investigated a variety of configurations for the above networks.
For all of our networks, we use CorrelateAll before the last linear layer during both training and testing.
This has the effect of not causing any loss of accuracy by that linear layer before the concretization of the loss function.

\textbf{None} means that no additional abstract layers are used.

\textbf{FewCombo} for ResNet-Tiny, has a CorrelateMax$_{32}$ layer before the first layer, a DecorrelateMin$_8$ after the first layer,
a DecorrelateMin$_4$ after the first wide residual block, a DecorrelateAll after the second wide residual block,
and a CorrelateMax$_{10}$ before the fully connected layers.

\textbf{ManyFixed} for ResNet-Tiny, has a CorrelateMax$_{32}$ layer before the first layer, a CorrelateFixed$_{16}$ then DecorrelateMin$_{16}$ after the first layer, a CorrelateFixed$_{8}$ then DecorrelateMin$_{8}$ after the first and second wide blocks, and a CorrelateFixed$_{4}$ then DecorrelateMin$_{4}$ after the third wide block, and a DecorrelateAll after the fourth.

\textbf{Combo} for SkipNet-18, has pairs of CorrelateFixed$_{k}$ and DecorrelateMin$_{\lfloor 0.5k \rfloor}$ with $k=20,10,5$ after layers 3, 4 and 5 respectively and
uses DeepLoss after the fourth layer with a weight schedule of Until(90,~Lin(0,~0.2,~50,~40),~0).

\textbf{LargeCombo} for ResNet-Large, has a CorrelateFixed$_{4}$ then DecorrelateMin$_{4}$ before wide residual blocks 1, 2, 3, and  4. Before the wide residual block 5, we place DecorrelateMin$_{2}$.   It uses DeepLoss after block 2 and 5, with weight schedules of Until(1,~0,~Lin(0.5,~0,~50,~3)) and Until(24,Lin(0,~0.1,~20,~4),~Lin(0.1,~0,~50)).

A complete description of each network with each abstract layer combination can be found in the Appendix, along with a table showing its performance for every training scheme.

\section{ Experimental Results }

\begin{figure*}[t]
\centering
\begin{subfigure}[b]{0.48\textwidth}
\centering
\begin{tikzpicture}[scale=0.8]\def\size{0.75}\begin{axis}[
  ybar = 11,
  scale=\size,
  width=12cm,
  height=8cm,
  xlabel={(a)},
  ylabel={Test Accuracy},
  symbolic x coords={Baseline, InSamp, InSampLPA, Adv$_3$InSampLPA},
  nodes near coords,
  xtick=data,
  yticklabel=$\pgfmathprintnumber{\tick}\%$,
  ymin=27, ymax=35,
  y axis line style={draw=none},
  axis background style = {fill=lightergrey},
  ylabel style={rotate=-90, at={(0.05,1.1)}, anchor=north west},
  y tick style={draw=none},
  legend cell align={left},
  legend style={draw=none, fill=none, at={(0.55,0.75)}, anchor=south west},
  xtick align=outside,
  axis y line*=left,
  axis x line*=bottom, 
  ymajorgrids=true,
  xmajorgrids=false,
  grid style={solid, draw=white},
]

\addplot[
  fill=orange,
  draw=none,
]
table[x=name, y=acc]
{plots/data/resnet_tiny.dat};

\addplot[
  fill=darkgreen!80,
  draw=none,
]
table[x=name, y=acc_abs_layers]
{plots/data/resnet_tiny.dat};

\legend{ No Abstract Layers
       , Abstract Layers
       }

\end{axis}\end{tikzpicture}
\end{subfigure}
\hspace{0.035\textwidth}%
\begin{subfigure}[b]{0.48\textwidth}
\centering
\begin{tikzpicture}[scale=0.8]\def\size{0.75}\begin{axis}[
  ybar = 11,
  scale=\size,
  width=12cm,
  height=8cm,
  xlabel={(b)},
  ylabel={Verified Robustness},
  symbolic x coords={Baseline, InSamp, InSampLPA, Adv$_3$InSampLPA},
  nodes near coords,
  xtick=data,
  yticklabel=$\pgfmathprintnumber{\tick}\%$,
  ymin=18.5, ymax=21.5,
  y axis line style={draw=none},
  axis background style = {fill=lightergrey},
  ylabel style={rotate=-90, at={(0.05,1.1)}, anchor=north west},
  y tick style={draw=none},
  legend cell align={left},
  legend style={draw=none, fill=none, at={(0.05,0.75)}, anchor=south west},
  xtick align=outside,
  axis y line*=left,
  axis x line*=bottom, 
  ymajorgrids=true,
  xmajorgrids=false,
  grid style={solid, draw=white},
]

\addplot[
  fill=orange,
  draw=none,
]
table[x=name, y=hbox]
{plots/data/resnet_tiny.dat};

\addplot[
  fill=darkgreen!80,
  draw=none,
]
table[x=name, y=hbox_abs_layers]
{plots/data/resnet_tiny.dat};

\end{axis}\end{tikzpicture}
\end{subfigure}
\vspace{-0.9cm}
\caption{A toy comparison of Accuracy (a) and Verified Robustness (b) of training schemes with similar parameters to Baseline on ResNet-Tiny with and without abstract layers. }
\label{resnet_tiny_graph}
\vspace{-0.6cm}
\end{figure*}
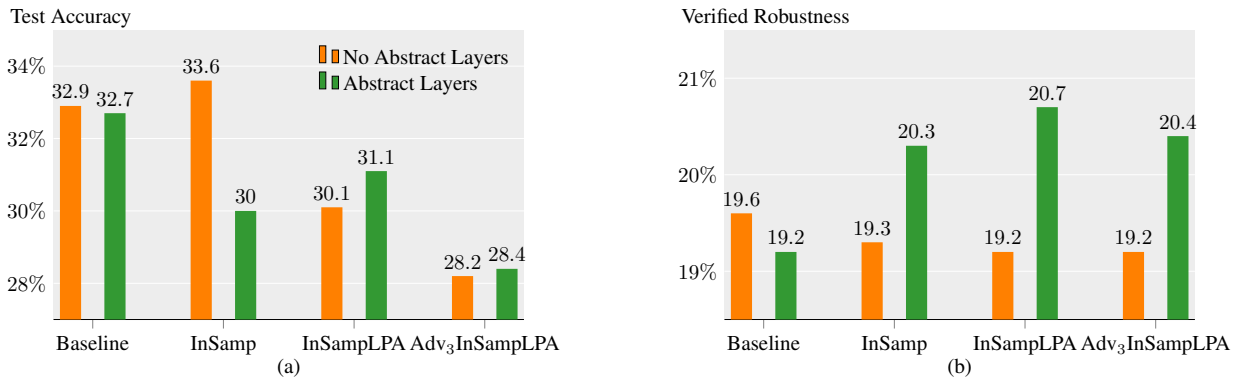

We now demonstrate how our training schemes shown in Table~\ref{tab:methods} (and discussed earlier) can be used to train provably robust networks of sizes an order of magnitude larger than prior work. We additionally show how abstract layers can be used to further push the envelope of provable robustness.

\paragraph{ Comparing Training Schemes and Abstract Layers }

To evaluate which combinations of training schemes and abstract layers provide the best results, we first trained ResNet-Tiny using four training schemes both without abstract layers, and with the abstract layer setup described by FewCombo. We trained each for 400 epochs.  The complete results are included in the Appendix, here we show the accuracy and verified robustness in Figure~\ref{resnet_tiny_graph}. We can observe that using abstract layers improves both provable robustness and accuracy when a more complex training scheme is used, and that benefits exist for  provable robustness as well. When the objective is only to maximize provable robustness, the training schemes (there are several) which utilize InSamp \emph{and} abstract layers, are optimal. Without abstract layers, inclusion sampling alone appears to have a benefit on accuracy without significant detriment to provable robustness.

To further investigate the effect of abstract layers, we compare the results of training three configurations of abstract layers on ResNet-Tiny.
These can be seen in Table~\ref{tab:tinyCompareAbsLayers}, which shows the results on the test set after training with Adv$_1$ISLPA for 350 epochs.

\begin{table}[]
\centering
\caption{ Comparison of abstract layers on ResNet-Tiny. }
\vskip 0.1in
{ \footnotesize
\begin{tabu}{@{}rrrrr@{}} \toprule
  Layers
& s/epoch
& Acc\%
& Attck\%
& Ver\%
\\ \midrule
None      & 130 & 29.4 & 21.4 & 17.7 \\
FewCombo  & 220 & 29.0 & 21.9 & 19.6 \\
ManyFixed & 345 & 28.9 & 21.4 & 19.2 \\
\bottomrule
\end{tabu}}
\label{tab:tinyCompareAbsLayers}
\vspace{-0.5cm}
\end{table}

In this experiment, we can see that ManyFixed actually does not perform as well as FewCombo in any metric, while for verified robustness both networks with abstract layers outperform the network without abstract layers.

While ManyFixed contains many more abstract layers and uses more correlation (thus making it significantly slower to train), the layers in FewCombo have been chosen more selectively. ManyFix contains multiple iterations of CorrelateFix$_k$ immediately before DecorrelateMin$_k$ of decreasing size.
We hypothesize that placing a CorrelateFix immediately before DecorrelateMin diminishes the utility of DecorrelateMin's heuristic.
As uncorrelated error coefficients tend to accumulate and grow while correlated error terms tend to shrink, a saddle point is generated wherein the network would need to maximize error coefficients (and thus decreasing accuracy) for neurons decided previously to be important (which will become correlated) and minimize error coefficients (and thus increasing accuracy) for neurons that will be decided to be unimportant in order to keep the coefficients from switching.

In summary, FewCombo is more efficient and more accurate than ManyFixed and is a key example for the necessity of the ``programming to prove'' methodology.

\paragraph{ Scaling to SkipNet-18 }

In order to build a defense scheme capable of training SkipNet-18 we found it necessary to, at a minimum, use InSamp$_{18}$ training.
Table~\ref{tab:resnet18compareMethod} demonstrates the result of training SkipNet-18 with a variety of training schemes for 400 epochs\footnote{ Baseline was stopped early at 350 epochs as it had clearly failed to train.}.
One can observe that Baseline diverged and while InSamp$_{18}$ without abstract layers or DeepLoss
was able to train a SkipNet-18 model with highest provable robustness, the highest accuracy was obtained using the same training scheme and Combo abstract layers.
The training scheme Adv$_5$IS$_{18}$ achieved a better compromise between provable robustness and accuracy.

\paragraph{ Larger And More Complex Architectures }
\begin{table}
\centering
\caption{ Comparison of training instances on SkipNet-18. }
\vskip 0.1in
{ \footnotesize
\begin{tabu}{@{}rrrrrr@{}} \toprule
  Scheme
& Layers
& s/epoch
& Acc\%
& Attck\%
& Ver\%
\\ \midrule
Baseline$_{18}$ & None & 152 & 10.2 & - & - \\
InSamp$_{18}$   & None & 102 & 28.5 & 23.4 & 20.5 \\
Adv$_5$IS$_{18}$ & None & 260 & 28.4 & 23.8 & 21.2 \\
InSamp$_{18}$   & Combo & 342 & 29.5 & 23 & 18.5 \\
\bottomrule
\end{tabu} }
\label{tab:resnet18compareMethod}
\vspace{-0.7cm}
\end{table}
While the majority of our comparisons were performed on ResNet-Tiny and SkipNet-18 we use our schemes to scale to significantly larger networks.
To show this, we designed a larger wide residual network, ResNet-Large, with 70k more neurons than SkipNet-18 and 66 million parameters (more than four times as many as SkipNet-18).
Here, we found it necessary to use a combination of previously evaluated techniques, in addition to two DeepLoss layers.

For this network we used the BiAdv$_L$ training scheme, which constructs abstract boxes from adversarial attacks.  
As training this network was significantly more expensive, taking 527 seconds per epoch, we halted training after 100 epochs. The results for this network can be seen in Table~\ref{tab:simple_nets}.  
While neither the accuracy nor verifiable robustness are particularly competitive with smaller networks, this is the deepest network (by shortest path from input to output) to have proved robust for a competitive epsilon value and that also comes with non-trivial accuracy.  While ResNet-34 and DenseNet-100 have longer paths from input to output, they also have very short paths which means that they could potentially learn a small and provably robust network first as an easier sub-problem.  

On deeper networks and larger networks that have more residual connections, we found that abstract layers were not as necessary for training. 
Here, we hypothesize that the network can provably learn the smaller network without the residual layers first, and then use them as possible when they do not too 
seriously hurt or provability.  Table~\ref{tab:simple_nets} also shows the results of training ResNet-18, ResNet-34, and DenseNet-100.  The largest, DenseNet-100 is 4.5 times the number of neurons to appear in any other paper at the time of this publication to have a non-trivial number of points 
verified to be robust.

\section{Conclusion}
We introduced a method for training provably robust networks based on the novel concept of abstract layers and a domain specific language for specifying complex training objectives. Our experimental evaluation demonstrates that our approach is effective in training provably robust networks that are an order of magnitude larger than those considered in prior work.

\nocite{cifar} 
\bibliography{paper}
\bibliographystyle{icml2019}

\ifdefined\ismain\else\clearpage
\appendix

\section{Evaluation}

\begin{strip}
\centering
\captionof{table}{ The networks compared and their sizes and speeds under different training schemes. }
\vskip 0.1in
{\footnotesize
\begin{tabu}{@{}rrrrrrr@{}} 
\toprule
Network Name
& Neurons
& Parameters
& Abstract Layers
& Training Scheme
& Batch Size
& Seconds Per Epoch
\\ \midrule
\multirow{11}{*}{ ResNet-Tiny } & \multirow{11}{*}{ 311796 } & \multirow{11}{*}{ 18415231 }
    & \multirow{5}{*}{ None }      & Baseline  & 50 & 106 \\ 
& & &                              & InSamp  & 50 & 106 \\ 
& & &                              & InSampLPA  & 50 & 107 \\ 
& & &                              & Adv$_1$ISLPA & 50 & 161 \\ 
& & &                              & Adv$_3$ISLPA & 50 & 130 \\ \clinelight{4-} 
& & & \multirow{5}{*}{ FewCombo }  & Baseline  & 50 & 209 \\ 
& & &                              & InSamp  & 50 & 205 \\ 
& & &                              & InSampLPA  & 50 & 206 \\ 
& & &                              & Adv$_1$ISLPA & 50 & 220 \\ 
& & &                              & Adv$_3$ISLPA & 50 & 265 \\ \clinelight{4-} 
& & & ManyFixed                    & Adv$_1$ISLPA & 50 & 347 \\ 
\clinelight{} 
\multirow{5}{*}{ SkipNet-18 } & \multirow{5}{*}{ 558080 } & \multirow{5}{*}{ 15626634 }
    & \multirow{3}{*}{ None }      & Baseline$_{18}$   & 200 & 152  \\
& & &                              & InSamp$_{18}$     & 200 & 102  \\
& & &                              & Adv$_5$IS$_{18}$ & 200 & 260  \\ \clinelight{4-} 
& & & Combo                        & InSamp$_{18}$     & 100 & 342 \\  \clinelight{} 
ResNet-Large & 639976 & 65819474   & LargeCombo      & BiAdv$_{L}$   & 50 & 527  \\ \clinelight{} 
ResNet-Large & 558k & 18m   & ResNet-18      & Adv$_5$ISLPA$_{R18}$   & 200 & 233  \\ \clinelight{} 
ResNet-Large & 967k & 25m   & ResNet-34      & InSampLPA$_{R34}$     & 200 & 176  \\ 
\bottomrule
\end{tabu} }
\label{tab:nets}
\end{strip}

\quad

\begin{strip}
\centering
\captionof{table}{ Comparison of different abstract layers and training schemes on ResNet-Tiny. }
\vskip 0.1in
{ \footnotesize
\begin{tabu}{@{}rrrrrr@{}} \toprule
  Train Scheme
& Abstract Layers
& Seconds Per Epoch
& Standard Accuracy \%
& Attacked Accuracy \%
& Verified Robust \%
\\ \midrule
\multirow{2}{*}{Baseline} & None     & 105 & 32.9 & 23.7 & 19.6 \\
                          & FewCombo & 209 & 32.7 & 24.1 & 19.2 \\ \clinelight{1-} 
\multirow{2}{*}{InSamp} & None     & 106 & 33.6 & 24.7 & 19.3  \\
                        & FewCombo & 205 & 30 & 23.2 & 20.3 \\ \clinelight{1-} 
\multirow{2}{*}{InSampLPA} & None & 107 & 30.1 & 22.5 & 19.2 \\
                           & FewCombo & 206 & 31.1 & 23 & 20.7 \\ \clinelight{1-} 
\multirow{2}{*}{Adv$_3$ISLPA} & None    & 161 & 28.2 & 22.2 & 19.2 \\
                                  & FewCombo & 267 & 28.4 & 22.5 & 20.4 \\
\bottomrule
\end{tabu} }
\label{tab:tinyCompareMethod}
\end{strip}

\quad

\clearpage
\section{Further Accuracy Results for MNIST}

\begin{strip}
\centering
\captionof{table}{ Networks for MNIST. }
\vskip 0.1in
{ \footnotesize
\begin{tabu}{@{}rrrr@{}} \toprule
Network Name
& Neurons
& Parameters
& Depth (ReLUs)
\\ \midrule
FFNN & 500 & 119910 & 5 \\
ConvSmall & 3604 & 89606 & 3 \\
ConvMed & 4804 & 166406 & 3 \\
ConvBig & 48064 & 1974762 & 6 \\
ConvLargeIBP & 175816 & 5426402 & 6 \\
TruncatedVGG & 151040 & 13109706 & 5  \\
\bottomrule
\end{tabu} }
\label{tab:mnist_nets}
\end{strip}
\begin{strip}
\centering
\captionof{table}{ MNIST with 0.1 }
\vskip 0.1in
{ \footnotesize
\begin{tabu}{@{}rrrr@{}} \toprule
Network & Standard Accuracy & PGD Accuracy & HBox Provability \\ \midrule
FFNN & 93.3\% & 90.8\% & 88.9\% \\ 
ConvSmall & 97.8\% & 96.2\% & 95.5\% \\
ConvMed & 97.8\% & 96.3\% & 95.5\% \\ 
ConvBig & 98.5\% & 97.2\% & 95.6\% \\
ConvLargeIBP & 98.7\% & 97.5\% & 95.8\% \\
TruncatedVGG & 98.9\% & 97.7\% & 95.6\% \\ 
\bottomrule
\end{tabu} }
\label{tab:mnist1}
\end{strip}
\begin{strip}
\centering
\captionof{table}{ MNIST with $\epsilon=0.3$. }
\vskip 0.1in
{ \footnotesize
\begin{tabu}{@{}rrrr@{}} \toprule
Network & Standard Accuracy & PGD Accuracy & HBox Provability \\ \midrule
FFNN & 80.2\% & 73.4\% & 62.6\% \\
ConvSmall & 96.9\% & 93.6\% & 89.1\% \\
ConvMed & 96.6\% & 93.1\% & 89.3\% \\
ConvBig & 97.0\% & 95.2\% & 87.8\% \\
ConvLargeIBP & 97.2\% & 95.4\% & 88.8\% \\
TruncatedVGG & 96.5\% & 94.4\% & 87.6\% \\
\bottomrule
\end{tabu} }
\label{tab:mnist3}
\end{strip}

\quad
\clearpage
\section{Further Accuracy Results for CIFAR10}

\begin{strip}
\centering
\captionof{table}{ Networks for CIFAR10. }
\vskip 0.1in
{ \footnotesize
\begin{tabu}{@{}rrrr@{}} \toprule
Network Name
& Neurons
& Parameters
& Depth (ReLUs)
\\ \midrule
FFNN & 500 & 348710 & 5 \\
ConvSmall & 4852 & 125318 & 3 \\
ConvMed & 6244 & 214918 & 3 \\
ConvBig & 62464 & 2466858 & 6 \\
ConvLargeIBP & 229576 & 6963554 & 6 \\
TruncatedVGG & 197120 & 17043018 & 5 \\
\bottomrule
\end{tabu} }
\label{tab:cifar_nets}
\end{strip}
\begin{strip}
\centering
\captionof{table}{ CIFAR10 with $\epsilon=3/255$. }
\vskip 0.1in
{ \footnotesize
\begin{tabu}{@{}rrrr@{}} \toprule
Network & Standard Accuracy & PGD Accuracy & HBox Provability \\ \midrule
FFNN& 45.1\% & 37.0\% & 33.1\% \\
ConvSmall& 56.1\% & 46.2\% & 42.4\% \\ 
ConvMed& 56.9\% & 46.6\% & 43.2\% \\ 
ConvBig& 61.9\% & 51.4\% & 45.0\% \\
ConvLargeIBP& 61.1\% & 51.4\% & 44.5\% \\ 
TruncatedVGG& 62.3\% & 51.4\% & 45.5\% \\ 
\bottomrule
\end{tabu} }
\label{tab:cifar3}
\end{strip}
\begin{strip}
\centering
\captionof{table}{ CIFAR10 with $\epsilon=8/255$. }
\vskip 0.1in
{ \footnotesize
\begin{tabu}{@{}rrrr@{}} \toprule
Network & Standard Accuracy & PGD Accuracy & HBox Provability \\ \midrule
FFNN& 33.5\% & 23.8\% & 19.0\% \\ 
ConvSmall& 42.6\% & 30.5\% & 24.9\% \\ 
ConvMed& 43.6\% & 30.3\% & 24.7\% \\ 
ConvBig& 46.0\% & 34.2\% & 25.2\% \\ 
ConvLargeIBP& 46.2\% & 34.7\% & 27.2\% \\
TruncatedVGG& 45.9\% & 34.4\% & 27.0\% \\
\bottomrule
\end{tabu} }
\label{tab:cifar8}
\end{strip}

\quad
\clearpage

\section{Networks and Abstract Layers}

\RecustomVerbatimCommand{\VerbatimInput}{VerbatimInput}%
{fontsize=\scriptsize,
}

\label{nets}

\begin{strip}
\captionof{figure}{ResNet-Tiny, None}
\begin{Verbatim}
Normalize mean=[0.4914, 0.4822, 0.4465] std=[0.2023, 0.1994, 0.201]
Conv2D, filters=16, kernel_size=[3, 3], input_shape=[32, 32, 3], stride=[1, 1], padding=1
ReLU
ParNet1
    Conv2D, filters=16, kernel_size=[1, 1], input_shape=[32, 32, 16], stride=[1, 1], padding=0
ParNet2
    Conv2D, filters=16, kernel_size=[3, 3], input_shape=[32, 32, 16], stride=[1, 1], padding=1
    ReLU
    Conv2D, filters=16, kernel_size=[3, 3], input_shape=[32, 32, 16], stride=[1, 1], padding=1
ParSum
ReLU
ParNet1
    Conv2D, filters=32, kernel_size=[1, 1], input_shape=[32, 32, 16], stride=[1, 1], padding=0
ParNet2
    Conv2D, filters=32, kernel_size=[3, 3], input_shape=[32, 32, 16], stride=[1, 1], padding=1
    ReLU
    Conv2D, filters=32, kernel_size=[3, 3], input_shape=[32, 32, 32], stride=[1, 1], padding=1
ParSum
ReLU
ParNet1
    Conv2D, filters=32, kernel_size=[1, 1], input_shape=[32, 32, 32], stride=[1, 1], padding=0
ParNet2
    Conv2D, filters=32, kernel_size=[3, 3], input_shape=[32, 32, 32], stride=[1, 1], padding=1
    ReLU
    Conv2D, filters=32, kernel_size=[3, 3], input_shape=[32, 32, 32], stride=[1, 1], padding=1
ParSum
ReLU
ParNet1
    Conv2D, filters=32, kernel_size=[1, 1], input_shape=[32, 32, 32], stride=[1, 1], padding=0
ParNet2
    Conv2D, filters=32, kernel_size=[3, 3], input_shape=[32, 32, 32], stride=[1, 1], padding=1
    ReLU
    Conv2D, filters=32, kernel_size=[3, 3], input_shape=[32, 32, 32], stride=[1, 1], padding=1
ParSum
ReLU
ParNet1
    Conv2D, filters=32, kernel_size=[1, 1], input_shape=[32, 32, 32], stride=[1, 1], padding=0
ParNet2
    Conv2D, filters=32, kernel_size=[3, 3], input_shape=[32, 32, 32], stride=[1, 1], padding=1
    ReLU
    Conv2D, filters=32, kernel_size=[3, 3], input_shape=[32, 32, 32], stride=[1, 1], padding=1
ParSum
ReLU
Linear out=500
ReLU
CorrelateAll only_train=False
Linear out=10
\end{Verbatim}
\end{strip}
\quad
\clearpage

\begin{strip}
\captionof{figure}{ResNet-Tiny,  FewCombo}
\begin{Verbatim}
Normalize mean=[0.4914, 0.4822, 0.4465] std=[0.2023, 0.1994, 0.201]
CorrMaxK only_train=True num_correlate=32
Conv2D, filters=16, kernel_size=[3, 3], input_shape=[32, 32, 3], stride=[1, 1], padding=1
ReLU
DecorrMin only_train=True k=8 num_to_keep=True
ParNet1
    Conv2D, filters=16, kernel_size=[1, 1], input_shape=[32, 32, 16], stride=[1, 1], padding=0
ParNet2
    Conv2D, filters=16, kernel_size=[3, 3], input_shape=[32, 32, 16], stride=[1, 1], padding=1
    ReLU
    Conv2D, filters=16, kernel_size=[3, 3], input_shape=[32, 32, 16], stride=[1, 1], padding=1
ParSum
ReLU
DecorrMin only_train=True k=4 num_to_keep=True
ParNet1
    Conv2D, filters=32, kernel_size=[1, 1], input_shape=[32, 32, 16], stride=[1, 1], padding=0
ParNet2
    Conv2D, filters=32, kernel_size=[3, 3], input_shape=[32, 32, 16], stride=[1, 1], padding=1
    ReLU
    Conv2D, filters=32, kernel_size=[3, 3], input_shape=[32, 32, 32], stride=[1, 1], padding=1
ParSum
ReLU
Concretize only_train=True
ParNet1
    Conv2D, filters=32, kernel_size=[1, 1], input_shape=[32, 32, 32], stride=[1, 1], padding=0
ParNet2
    Conv2D, filters=32, kernel_size=[3, 3], input_shape=[32, 32, 32], stride=[1, 1], padding=1
    ReLU
    Conv2D, filters=32, kernel_size=[3, 3], input_shape=[32, 32, 32], stride=[1, 1], padding=1
ParSum
ReLU
ParNet1
    Conv2D, filters=32, kernel_size=[1, 1], input_shape=[32, 32, 32], stride=[1, 1], padding=0
ParNet2
    Conv2D, filters=32, kernel_size=[3, 3], input_shape=[32, 32, 32], stride=[1, 1], padding=1
    ReLU
    Conv2D, filters=32, kernel_size=[3, 3], input_shape=[32, 32, 32], stride=[1, 1], padding=1
ParSum
ReLU
ParNet1
    Conv2D, filters=32, kernel_size=[1, 1], input_shape=[32, 32, 32], stride=[1, 1], padding=0
ParNet2
    Conv2D, filters=32, kernel_size=[3, 3], input_shape=[32, 32, 32], stride=[1, 1], padding=1
    ReLU
    Conv2D, filters=32, kernel_size=[3, 3], input_shape=[32, 32, 32], stride=[1, 1], padding=1
ParSum
ReLU
CorrMaxK only_train=True num_correlate=10
Linear out=500
ReLU
CorrelateAll only_train=False
Linear out=10
\end{Verbatim}
\end{strip}
\quad
\clearpage

\begin{strip}
\captionof{figure}{ResNet-Tiny, ManyFixed}
\begin{Verbatim}
Normalize mean=[0.4914, 0.4822, 0.4465] std=[0.2023, 0.1994, 0.201]
CorrMaxK only_train=True num_correlate=32
Conv2D, filters=16, kernel_size=[3, 3], input_shape=[32, 32, 3], stride=[1, 1], padding=1
ReLU
CorrFix only_train=True k=16
DecorrMin only_train=True k=16 num_to_keep=True
ParNet1
    Conv2D, filters=16, kernel_size=[1, 1], input_shape=[32, 32, 16], stride=[1, 1], padding=0
ParNet2
    Conv2D, filters=16, kernel_size=[3, 3], input_shape=[32, 32, 16], stride=[1, 1], padding=1
    ReLU
    Conv2D, filters=16, kernel_size=[3, 3], input_shape=[32, 32, 16], stride=[1, 1], padding=1
ParSum
ReLU
CorrFix only_train=True k=8
DecorrMin only_train=True k=8 num_to_keep=True
ParNet1
    Conv2D, filters=32, kernel_size=[1, 1], input_shape=[32, 32, 16], stride=[1, 1], padding=0
ParNet2
    Conv2D, filters=32, kernel_size=[3, 3], input_shape=[32, 32, 16], stride=[1, 1], padding=1
    ReLU
    Conv2D, filters=32, kernel_size=[3, 3], input_shape=[32, 32, 32], stride=[1, 1], padding=1
ParSum
ReLU
CorrFix only_train=True k=8
DecorrMin only_train=True k=8 num_to_keep=True
ParNet1
    Conv2D, filters=32, kernel_size=[1, 1], input_shape=[32, 32, 32], stride=[1, 1], padding=0
ParNet2
    Conv2D, filters=32, kernel_size=[3, 3], input_shape=[32, 32, 32], stride=[1, 1], padding=1
    ReLU
    Conv2D, filters=32, kernel_size=[3, 3], input_shape=[32, 32, 32], stride=[1, 1], padding=1
ParSum
ReLU
CorrFix only_train=True k=4
DecorrMin only_train=True k=4 num_to_keep=True
ParNet1
    Conv2D, filters=32, kernel_size=[1, 1], input_shape=[32, 32, 32], stride=[1, 1], padding=0
ParNet2
    Conv2D, filters=32, kernel_size=[3, 3], input_shape=[32, 32, 32], stride=[1, 1], padding=1
    ReLU
    Conv2D, filters=32, kernel_size=[3, 3], input_shape=[32, 32, 32], stride=[1, 1], padding=1
ParSum
ReLU
Concretize only_train=True
ParNet1
    Conv2D, filters=32, kernel_size=[1, 1], input_shape=[32, 32, 32], stride=[1, 1], padding=0
ParNet2
    Conv2D, filters=32, kernel_size=[3, 3], input_shape=[32, 32, 32], stride=[1, 1], padding=1
    ReLU
    Conv2D, filters=32, kernel_size=[3, 3], input_shape=[32, 32, 32], stride=[1, 1], padding=1
ParSum
ReLU
Linear out=500
ReLU
CorrelateAll only_train=False
Linear out=10
\end{Verbatim}
\end{strip}
\quad
\clearpage

\begin{strip}
\captionof{figure}{SkipNet-18, None}
\begin{Verbatim}
Normalize mean=[0.4914, 0.4822, 0.4465] std=[0.2023, 0.1994, 0.201]
Conv2D, filters=64, kernel_size=[3, 3], input_shape=[32, 32, 3], stride=[1, 1], padding=1
ReLU
Conv2D, filters=64, kernel_size=[3, 3], input_shape=[32, 32, 64], stride=[1, 1], padding=1
ReLU
Conv2D, filters=64, kernel_size=[3, 3], input_shape=[32, 32, 64], stride=[1, 1], padding=1
ReLU
Conv2D, filters=64, kernel_size=[3, 3], input_shape=[32, 32, 64], stride=[1, 1], padding=1
ReLU
Conv2D, filters=64, kernel_size=[3, 3], input_shape=[32, 32, 64], stride=[1, 1], padding=1
ReLU
ParNet1
    Conv2D, filters=128, kernel_size=[3, 3], input_shape=[32, 32, 64], stride=[2, 2], padding=1
    ReLU
    Conv2D, filters=128, kernel_size=[3, 3], input_shape=[16, 16, 128], stride=[1, 1], padding=1
ParNet2
    Conv2D, filters=128, kernel_size=[1, 1], input_shape=[32, 32, 64], stride=[2, 2], padding=0
ParSum
ReLU
Conv2D, filters=128, kernel_size=[3, 3], input_shape=[16, 16, 128], stride=[1, 1], padding=1
ReLU
Conv2D, filters=128, kernel_size=[3, 3], input_shape=[16, 16, 128], stride=[1, 1], padding=1
ReLU
ParNet1
    Conv2D, filters=256, kernel_size=[3, 3], input_shape=[16, 16, 128], stride=[2, 2], padding=1
    ReLU
    Conv2D, filters=256, kernel_size=[3, 3], input_shape=[8, 8, 256], stride=[1, 1], padding=1
ParNet2
    Conv2D, filters=256, kernel_size=[1, 1], input_shape=[16, 16, 128], stride=[2, 2], padding=0
ParSum
ReLU
Conv2D, filters=256, kernel_size=[3, 3], input_shape=[8, 8, 256], stride=[1, 1], padding=1
ReLU
Conv2D, filters=256, kernel_size=[3, 3], input_shape=[8, 8, 256], stride=[1, 1], padding=1
ReLU
ParNet1
    Conv2D, filters=512, kernel_size=[3, 3], input_shape=[8, 8, 256], stride=[2, 2], padding=1
    ReLU
    Conv2D, filters=512, kernel_size=[3, 3], input_shape=[4, 4, 512], stride=[1, 1], padding=1
ParNet2
    Conv2D, filters=512, kernel_size=[1, 1], input_shape=[8, 8, 256], stride=[2, 2], padding=0
ParSum
ReLU
Conv2D, filters=512, kernel_size=[3, 3], input_shape=[4, 4, 512], stride=[1, 1], padding=1
ReLU
Conv2D, filters=512, kernel_size=[3, 3], input_shape=[4, 4, 512], stride=[1, 1], padding=1
ReLU
Linear out=512
ReLU
Linear out=512
ReLU
CorrelateAll only_train=False
Linear out=10
\end{Verbatim}
\end{strip}
\quad
\clearpage

\begin{strip}
\captionof{figure}{SkipNet-18, Combo}
\begin{Verbatim}
Normalize mean=[0.4914, 0.4822, 0.4465] std=[0.2023, 0.1994, 0.201]
Conv2D, filters=64, kernel_size=[3, 3], input_shape=[32, 32, 3], stride=[1, 1], padding=1
ReLU
Conv2D, filters=64, kernel_size=[3, 3], input_shape=[32, 32, 64], stride=[1, 1], padding=1
ReLU
Conv2D, filters=64, kernel_size=[3, 3], input_shape=[32, 32, 64], stride=[1, 1], padding=1
ReLU
Conv2D, filters=64, kernel_size=[3, 3], input_shape=[32, 32, 64], stride=[1, 1], padding=1
ReLU
Conv2D, filters=64, kernel_size=[3, 3], input_shape=[32, 32, 64], stride=[1, 1], padding=1
ReLU
ParNet1
    Conv2D, filters=128, kernel_size=[3, 3], input_shape=[32, 32, 64], stride=[2, 2], padding=1
    ReLU
    Conv2D, filters=128, kernel_size=[3, 3], input_shape=[16, 16, 128], stride=[1, 1], padding=1
ParNet2
    Conv2D, filters=128, kernel_size=[1, 1], input_shape=[32, 32, 64], stride=[2, 2], padding=0
ParSum
ReLU
CorrFix only_train=True k=20
DecorrMin only_train=True k=10 num_to_keep=True
Conv2D, filters=128, kernel_size=[3, 3], input_shape=[16, 16, 128], stride=[1, 1], padding=1
ReLU
Conv2D, filters=128, kernel_size=[3, 3], input_shape=[16, 16, 128], stride=[1, 1], padding=1
ReLU
CorrFix only_train=True k=10
DecorrMin only_train=True k=5 num_to_keep=True
DeepLoss only_train=True bw=Until(90, Lin(%s,%s,%s,%s), 0.0) act=<function relu at 0x10b94d620>
ParNet1
    Conv2D, filters=256, kernel_size=[3, 3], input_shape=[16, 16, 128], stride=[2, 2], padding=1
    ReLU
    Conv2D, filters=256, kernel_size=[3, 3], input_shape=[8, 8, 256], stride=[1, 1], padding=1
ParNet2
    Conv2D, filters=256, kernel_size=[1, 1], input_shape=[16, 16, 128], stride=[2, 2], padding=0
ParSum
ReLU
CorrFix only_train=True k=5
DecorrMin only_train=True k=2 num_to_keep=True
Conv2D, filters=256, kernel_size=[3, 3], input_shape=[8, 8, 256], stride=[1, 1], padding=1
ReLU
Conv2D, filters=256, kernel_size=[3, 3], input_shape=[8, 8, 256], stride=[1, 1], padding=1
ReLU
ParNet1
    Conv2D, filters=512, kernel_size=[3, 3], input_shape=[8, 8, 256], stride=[2, 2], padding=1
    ReLU
    Conv2D, filters=512, kernel_size=[3, 3], input_shape=[4, 4, 512], stride=[1, 1], padding=1
ParNet2
    Conv2D, filters=512, kernel_size=[1, 1], input_shape=[8, 8, 256], stride=[2, 2], padding=0
ParSum
ReLU
Conv2D, filters=512, kernel_size=[3, 3], input_shape=[4, 4, 512], stride=[1, 1], padding=1
ReLU
Conv2D, filters=512, kernel_size=[3, 3], input_shape=[4, 4, 512], stride=[1, 1], padding=1
ReLU
Linear out=512
ReLU
Linear out=512
ReLU
CorrelateAll only_train=False
Linear out=10
\end{Verbatim}
\end{strip}
\quad
\clearpage

\begin{strip}
\captionof{figure}{ResNet-Large, LargeCombo}
\begin{Verbatim}
Normalize mean=[0.4914, 0.4822, 0.4465] std=[0.2023, 0.1994, 0.201]
Conv2D, filters=16, kernel_size=[3, 3], input_shape=[32, 32, 3], stride=[1, 1], padding=1
ReLU
CorrMaxK only_train=True num_correlate=4
ParNet1
    Conv2D, filters=16, kernel_size=[1, 1], input_shape=[32, 32, 16], stride=[1, 1], padding=0
ParNet2
    Conv2D, filters=16, kernel_size=[3, 3], input_shape=[32, 32, 16], stride=[1, 1], padding=1
    ReLU
    Conv2D, filters=16, kernel_size=[3, 3], input_shape=[32, 32, 16], stride=[1, 1], padding=1
ParSum
ReLU
CorrMaxK only_train=True num_correlate=4
DecorrMin only_train=True k=4 num_to_keep=True
ParNet1
    Conv2D, filters=32, kernel_size=[1, 1], input_shape=[32, 32, 16], stride=[1, 1], padding=0
ParNet2
    Conv2D, filters=32, kernel_size=[3, 3], input_shape=[32, 32, 16], stride=[1, 1], padding=1
    ReLU
    Conv2D, filters=32, kernel_size=[3, 3], input_shape=[32, 32, 32], stride=[1, 1], padding=1
ParSum
ReLU
CorrMaxK only_train=True num_correlate=4
DecorrMin only_train=True k=4 num_to_keep=True
ParNet1
    Conv2D, filters=32, kernel_size=[1, 1], input_shape=[32, 32, 32], stride=[1, 1], padding=0
ParNet2
    Conv2D, filters=32, kernel_size=[3, 3], input_shape=[32, 32, 32], stride=[1, 1], padding=1
    ReLU
    Conv2D, filters=32, kernel_size=[3, 3], input_shape=[32, 32, 32], stride=[1, 1], padding=1
ParSum
ReLU
DeepLoss only_train=True bw=Until(1, 0.0, Lin(0.5,0,50,3))
ParNet1
    Conv2D, filters=32, kernel_size=[1, 1], input_shape=[32, 32, 32], stride=[1, 1], padding=0
ParNet2
    Conv2D, filters=32, kernel_size=[3, 3], input_shape=[32, 32, 32], stride=[1, 1], padding=1
    ReLU
    Conv2D, filters=32, kernel_size=[3, 3], input_shape=[32, 32, 32], stride=[1, 1], padding=1
ParSum
ReLU
CorrMaxK only_train=True num_correlate=4
DecorrMin only_train=True k=4 num_to_keep=True
ParNet1
    Conv2D, filters=64, kernel_size=[1, 1], input_shape=[32, 32, 32], stride=[1, 1], padding=0
ParNet2
    Conv2D, filters=64, kernel_size=[3, 3], input_shape=[32, 32, 32], stride=[1, 1], padding=1
    ReLU
    Conv2D, filters=64, kernel_size=[3, 3], input_shape=[32, 32, 64], stride=[1, 1], padding=1
ParSum
ReLU
CorrMaxK only_train=True num_correlate=4
DecorrMin only_train=True k=2 num_to_keep=True
ParNet1
    Conv2D, filters=64, kernel_size=[1, 1], input_shape=[32, 32, 64], stride=[1, 1], padding=0
ParNet2
    Conv2D, filters=64, kernel_size=[3, 3], input_shape=[32, 32, 64], stride=[1, 1], padding=1
    ReLU
    Conv2D, filters=64, kernel_size=[3, 3], input_shape=[32, 32, 64], stride=[1, 1], padding=1
ParSum
ReLU
DeepLoss only_train=True bw=Until(24, Lin(0,0.1,20,4), Lin(0.1,0,50,0))
ParNet1
    Conv2D, filters=64, kernel_size=[1, 1], input_shape=[32, 32, 64], stride=[1, 1], padding=0
ParNet2
    Conv2D, filters=64, kernel_size=[3, 3], input_shape=[32, 32, 64], stride=[1, 1], padding=1
    ReLU
    Conv2D, filters=64, kernel_size=[3, 3], input_shape=[32, 32, 64], stride=[1, 1], padding=1
ParSum
ReLU
Linear out=1000
ReLU
CorrelateAll only_train=False
Linear out=10
\end{Verbatim}
\end{strip}
\quad
\clearpage

\begin{strip}
\captionof{figure}{ResNet-18, None}
\begin{Verbatim}
Normalize mean=[0.4914, 0.4822, 0.4465] std=[0.2023, 0.1994, 0.201]
Conv2D, filters=64, kernel_size=[3, 3], input_shape=[32, 32, 3], stride=[1, 1], padding=1
ReLU
ParNet1
    Conv2D, filters=64, kernel_size=[3, 3], input_shape=[32, 32, 64], stride=[1, 1], padding=1
    ReLU
    Conv2D, filters=64, kernel_size=[3, 3], input_shape=[32, 32, 64], stride=[1, 1], padding=1
ParNet2
ParSum
ReLU
ParNet1
    Conv2D, filters=64, kernel_size=[3, 3], input_shape=[32, 32, 64], stride=[1, 1], padding=1
    ReLU
    Conv2D, filters=64, kernel_size=[3, 3], input_shape=[32, 32, 64], stride=[1, 1], padding=1
ParNet2
ParSum
ReLU
ParNet1
    Conv2D, filters=128, kernel_size=[3, 3], input_shape=[32, 32, 64], stride=[2, 2], padding=1
    ReLU
    Conv2D, filters=128, kernel_size=[3, 3], input_shape=[16, 16, 128], stride=[1, 1], padding=1
ParNet2
    Conv2D, filters=128, kernel_size=[1, 1], input_shape=[32, 32, 64], stride=[2, 2], padding=0
ParSum
ReLU
ParNet1
    Conv2D, filters=128, kernel_size=[3, 3], input_shape=[16, 16, 128], stride=[1, 1], padding=1
    ReLU
    Conv2D, filters=128, kernel_size=[3, 3], input_shape=[16, 16, 128], stride=[1, 1], padding=1
ParNet2
ParSum
ReLU
ParNet1
    Conv2D, filters=256, kernel_size=[3, 3], input_shape=[16, 16, 128], stride=[2, 2], padding=1
    ReLU
    Conv2D, filters=256, kernel_size=[3, 3], input_shape=[8, 8, 256], stride=[1, 1], padding=1
ParNet2
    Conv2D, filters=256, kernel_size=[1, 1], input_shape=[16, 16, 128], stride=[2, 2], padding=0
ParSum
ReLU
ParNet1
    Conv2D, filters=256, kernel_size=[3, 3], input_shape=[8, 8, 256], stride=[1, 1], padding=1
    ReLU
    Conv2D, filters=256, kernel_size=[3, 3], input_shape=[8, 8, 256], stride=[1, 1], padding=1
ParNet2
ParSum
ReLU
ParNet1
    Conv2D, filters=512, kernel_size=[3, 3], input_shape=[8, 8, 256], stride=[2, 2], padding=1
    ReLU
    Conv2D, filters=512, kernel_size=[3, 3], input_shape=[4, 4, 512], stride=[1, 1], padding=1
ParNet2
    Conv2D, filters=512, kernel_size=[1, 1], input_shape=[8, 8, 256], stride=[2, 2], padding=0
ParSum
ReLU
ParNet1
    Conv2D, filters=512, kernel_size=[3, 3], input_shape=[4, 4, 512], stride=[1, 1], padding=1
    ReLU
    Conv2D, filters=512, kernel_size=[3, 3], input_shape=[4, 4, 512], stride=[1, 1], padding=1
ParNet2
ParSum
ReLU
Linear out=512
ReLU
Linear out=512
ReLU
CorrelateAll only_train=False
Linear out=10

\end{Verbatim}
\end{strip}
\quad
\clearpage

\begin{strip}
\captionof{figure}{ResNet-34, None}
\begin{Verbatim}
Normalize mean=[0.4914, 0.4822, 0.4465] std=[0.2023, 0.1994, 0.201]
Conv2D, filters=64, kernel_size=[3, 3], input_shape=[32, 32, 3], stride=[1, 1], padding=1
ReLU
ParNet1
    Conv2D, filters=64, kernel_size=[3, 3], input_shape=[32, 32, 64], stride=[1, 1], padding=1
    ReLU
    Conv2D, filters=64, kernel_size=[3, 3], input_shape=[32, 32, 64], stride=[1, 1], padding=1
ParNet2
ParSum
ReLU
ParNet1
    Conv2D, filters=64, kernel_size=[3, 3], input_shape=[32, 32, 64], stride=[1, 1], padding=1
    ReLU
    Conv2D, filters=64, kernel_size=[3, 3], input_shape=[32, 32, 64], stride=[1, 1], padding=1
ParNet2
ParSum
ReLU
ParNet1
    Conv2D, filters=64, kernel_size=[3, 3], input_shape=[32, 32, 64], stride=[1, 1], padding=1
    ReLU
    Conv2D, filters=64, kernel_size=[3, 3], input_shape=[32, 32, 64], stride=[1, 1], padding=1
ParNet2
ParSum
ReLU
ParNet1
    Conv2D, filters=128, kernel_size=[3, 3], input_shape=[32, 32, 64], stride=[2, 2], padding=1
    ReLU
    Conv2D, filters=128, kernel_size=[3, 3], input_shape=[16, 16, 128], stride=[1, 1], padding=1
ParNet2
    Conv2D, filters=128, kernel_size=[1, 1], input_shape=[32, 32, 64], stride=[2, 2], padding=0
ParSum
ReLU
ParNet1
    Conv2D, filters=128, kernel_size=[3, 3], input_shape=[16, 16, 128], stride=[1, 1], padding=1
    ReLU
    Conv2D, filters=128, kernel_size=[3, 3], input_shape=[16, 16, 128], stride=[1, 1], padding=1
ParNet2
ParSum
ReLU
ParNet1
    Conv2D, filters=128, kernel_size=[3, 3], input_shape=[16, 16, 128], stride=[1, 1], padding=1
    ReLU
    Conv2D, filters=128, kernel_size=[3, 3], input_shape=[16, 16, 128], stride=[1, 1], padding=1
ParNet2
ParSum
ReLU
ParNet1
    Conv2D, filters=128, kernel_size=[3, 3], input_shape=[16, 16, 128], stride=[1, 1], padding=1
    ReLU
    Conv2D, filters=128, kernel_size=[3, 3], input_shape=[16, 16, 128], stride=[1, 1], padding=1
ParNet2
ParSum
ReLU
ParNet1
    Conv2D, filters=256, kernel_size=[3, 3], input_shape=[16, 16, 128], stride=[2, 2], padding=1
    ReLU
    Conv2D, filters=256, kernel_size=[3, 3], input_shape=[8, 8, 256], stride=[1, 1], padding=1
ParNet2
    Conv2D, filters=256, kernel_size=[1, 1], input_shape=[16, 16, 128], stride=[2, 2], padding=0
ParSum
ReLU
ParNet1
    Conv2D, filters=256, kernel_size=[3, 3], input_shape=[8, 8, 256], stride=[1, 1], padding=1
    ReLU
    Conv2D, filters=256, kernel_size=[3, 3], input_shape=[8, 8, 256], stride=[1, 1], padding=1
ParNet2
ParSum
ReLU
ParNet1
    Conv2D, filters=256, kernel_size=[3, 3], input_shape=[8, 8, 256], stride=[1, 1], padding=1
    ReLU
    Conv2D, filters=256, kernel_size=[3, 3], input_shape=[8, 8, 256], stride=[1, 1], padding=1
ParNet2
ParSum
ReLU
ParNet1
    Conv2D, filters=256, kernel_size=[3, 3], input_shape=[8, 8, 256], stride=[1, 1], padding=1
    ReLU
    Conv2D, filters=256, kernel_size=[3, 3], input_shape=[8, 8, 256], stride=[1, 1], padding=1
ParNet2
ParSum
ReLU
ParNet1
    Conv2D, filters=256, kernel_size=[3, 3], input_shape=[8, 8, 256], stride=[1, 1], padding=1
    ReLU
    Conv2D, filters=256, kernel_size=[3, 3], input_shape=[8, 8, 256], stride=[1, 1], padding=1
ParNet2
ParSum
ReLU
ParNet1
    Conv2D, filters=256, kernel_size=[3, 3], input_shape=[8, 8, 256], stride=[1, 1], padding=1
    ReLU
    Conv2D, filters=256, kernel_size=[3, 3], input_shape=[8, 8, 256], stride=[1, 1], padding=1
ParNet2
ParSum
ReLU
ParNet1
    Conv2D, filters=512, kernel_size=[3, 3], input_shape=[8, 8, 256], stride=[2, 2], padding=1
    ReLU
    Conv2D, filters=512, kernel_size=[3, 3], input_shape=[4, 4, 512], stride=[1, 1], padding=1
ParNet2
    Conv2D, filters=512, kernel_size=[1, 1], input_shape=[8, 8, 256], stride=[2, 2], padding=0
ParSum
ReLU
ParNet1
    Conv2D, filters=512, kernel_size=[3, 3], input_shape=[4, 4, 512], stride=[1, 1], padding=1
    ReLU
    Conv2D, filters=512, kernel_size=[3, 3], input_shape=[4, 4, 512], stride=[1, 1], padding=1
ParNet2
ParSum
ReLU
ParNet1
    Conv2D, filters=512, kernel_size=[3, 3], input_shape=[4, 4, 512], stride=[1, 1], padding=1
    ReLU
    Conv2D, filters=512, kernel_size=[3, 3], input_shape=[4, 4, 512], stride=[1, 1], padding=1
ParNet2
ParSum
ReLU
Linear out=512
ReLU
Linear out=512
ReLU
CorrelateAll only_train=False
Linear out=10
\end{Verbatim}
\end{strip}
\quad
\clearpage

\begin{strip}
\captionof{figure}{DenseNet-100, None}
\begin{Verbatim}
Normalize mean=[0.4914, 0.4822, 0.4465] std=[0.2023, 0.1994, 0.201]
Conv2D, filters=24, kernel_size=[3, 3], input_shape=[32, 32, 3], stride=[1, 1], padding=1
SkipNet1
SkipNet2
    ReLU
    Conv2D, filters=48, kernel_size=[1, 1], input_shape=[32, 32, 24], stride=[1, 1], padding=0
    ReLU
    Conv2D, filters=12, kernel_size=[3, 3], input_shape=[32, 32, 48], stride=[1, 1], padding=1
SkipCat dim=1
SkipNet1
SkipNet2
    ReLU
    Conv2D, filters=48, kernel_size=[1, 1], input_shape=[32, 32, 36], stride=[1, 1], padding=0
    ReLU
    Conv2D, filters=12, kernel_size=[3, 3], input_shape=[32, 32, 48], stride=[1, 1], padding=1
SkipCat dim=1
SkipNet1
SkipNet2
    ReLU
    Conv2D, filters=48, kernel_size=[1, 1], input_shape=[32, 32, 48], stride=[1, 1], padding=0
    ReLU
    Conv2D, filters=12, kernel_size=[3, 3], input_shape=[32, 32, 48], stride=[1, 1], padding=1
SkipCat dim=1
SkipNet1
SkipNet2
    ReLU
    Conv2D, filters=48, kernel_size=[1, 1], input_shape=[32, 32, 60], stride=[1, 1], padding=0
    ReLU
    Conv2D, filters=12, kernel_size=[3, 3], input_shape=[32, 32, 48], stride=[1, 1], padding=1
SkipCat dim=1
SkipNet1
SkipNet2
    ReLU
    Conv2D, filters=48, kernel_size=[1, 1], input_shape=[32, 32, 72], stride=[1, 1], padding=0
    ReLU
    Conv2D, filters=12, kernel_size=[3, 3], input_shape=[32, 32, 48], stride=[1, 1], padding=1
SkipCat dim=1
SkipNet1
SkipNet2
    ReLU
    Conv2D, filters=48, kernel_size=[1, 1], input_shape=[32, 32, 84], stride=[1, 1], padding=0
    ReLU
    Conv2D, filters=12, kernel_size=[3, 3], input_shape=[32, 32, 48], stride=[1, 1], padding=1
SkipCat dim=1
SkipNet1
SkipNet2
    ReLU
    Conv2D, filters=48, kernel_size=[1, 1], input_shape=[32, 32, 96], stride=[1, 1], padding=0
    ReLU
    Conv2D, filters=12, kernel_size=[3, 3], input_shape=[32, 32, 48], stride=[1, 1], padding=1
SkipCat dim=1
SkipNet1
SkipNet2
    ReLU
    Conv2D, filters=48, kernel_size=[1, 1], input_shape=[32, 32, 108], stride=[1, 1], padding=0
    ReLU
    Conv2D, filters=12, kernel_size=[3, 3], input_shape=[32, 32, 48], stride=[1, 1], padding=1
SkipCat dim=1
SkipNet1
SkipNet2
    ReLU
    Conv2D, filters=48, kernel_size=[1, 1], input_shape=[32, 32, 120], stride=[1, 1], padding=0
    ReLU
    Conv2D, filters=12, kernel_size=[3, 3], input_shape=[32, 32, 48], stride=[1, 1], padding=1
SkipCat dim=1
SkipNet1
SkipNet2
    ReLU
    Conv2D, filters=48, kernel_size=[1, 1], input_shape=[32, 32, 132], stride=[1, 1], padding=0
    ReLU
    Conv2D, filters=12, kernel_size=[3, 3], input_shape=[32, 32, 48], stride=[1, 1], padding=1
SkipCat dim=1
SkipNet1
SkipNet2
    ReLU
    Conv2D, filters=48, kernel_size=[1, 1], input_shape=[32, 32, 144], stride=[1, 1], padding=0
    ReLU
    Conv2D, filters=12, kernel_size=[3, 3], input_shape=[32, 32, 48], stride=[1, 1], padding=1
SkipCat dim=1
SkipNet1
SkipNet2
    ReLU
    Conv2D, filters=48, kernel_size=[1, 1], input_shape=[32, 32, 156], stride=[1, 1], padding=0
    ReLU
    Conv2D, filters=12, kernel_size=[3, 3], input_shape=[32, 32, 48], stride=[1, 1], padding=1
SkipCat dim=1
SkipNet1
SkipNet2
    ReLU
    Conv2D, filters=48, kernel_size=[1, 1], input_shape=[32, 32, 168], stride=[1, 1], padding=0
    ReLU
    Conv2D, filters=12, kernel_size=[3, 3], input_shape=[32, 32, 48], stride=[1, 1], padding=1
SkipCat dim=1
SkipNet1
SkipNet2
    ReLU
    Conv2D, filters=48, kernel_size=[1, 1], input_shape=[32, 32, 180], stride=[1, 1], padding=0
    ReLU
    Conv2D, filters=12, kernel_size=[3, 3], input_shape=[32, 32, 48], stride=[1, 1], padding=1
SkipCat dim=1
SkipNet1
SkipNet2
    ReLU
    Conv2D, filters=48, kernel_size=[1, 1], input_shape=[32, 32, 192], stride=[1, 1], padding=0
    ReLU
    Conv2D, filters=12, kernel_size=[3, 3], input_shape=[32, 32, 48], stride=[1, 1], padding=1
SkipCat dim=1
SkipNet1
SkipNet2
    ReLU
    Conv2D, filters=48, kernel_size=[1, 1], input_shape=[32, 32, 204], stride=[1, 1], padding=0
    ReLU
    Conv2D, filters=12, kernel_size=[3, 3], input_shape=[32, 32, 48], stride=[1, 1], padding=1
SkipCat dim=1
ReLU
Conv2D, filters=108, kernel_size=[1, 1], input_shape=[32, 32, 216], stride=[1, 1], padding=0
AvgPool2D
SkipNet1
SkipNet2
    ReLU
    Conv2D, filters=48, kernel_size=[1, 1], input_shape=[17, 17, 108], stride=[1, 1], padding=0
    ReLU
    Conv2D, filters=12, kernel_size=[3, 3], input_shape=[17, 17, 48], stride=[1, 1], padding=1
SkipCat dim=1
SkipNet1
SkipNet2
    ReLU
    Conv2D, filters=48, kernel_size=[1, 1], input_shape=[17, 17, 120], stride=[1, 1], padding=0
    ReLU
    Conv2D, filters=12, kernel_size=[3, 3], input_shape=[17, 17, 48], stride=[1, 1], padding=1
SkipCat dim=1
SkipNet1
SkipNet2
    ReLU
    Conv2D, filters=48, kernel_size=[1, 1], input_shape=[17, 17, 132], stride=[1, 1], padding=0
    ReLU
    Conv2D, filters=12, kernel_size=[3, 3], input_shape=[17, 17, 48], stride=[1, 1], padding=1
SkipCat dim=1
SkipNet1
SkipNet2
    ReLU
    Conv2D, filters=48, kernel_size=[1, 1], input_shape=[17, 17, 144], stride=[1, 1], padding=0
    ReLU
    Conv2D, filters=12, kernel_size=[3, 3], input_shape=[17, 17, 48], stride=[1, 1], padding=1
SkipCat dim=1
SkipNet1
SkipNet2
    ReLU
    Conv2D, filters=48, kernel_size=[1, 1], input_shape=[17, 17, 156], stride=[1, 1], padding=0
    ReLU
    Conv2D, filters=12, kernel_size=[3, 3], input_shape=[17, 17, 48], stride=[1, 1], padding=1
SkipCat dim=1
SkipNet1
SkipNet2
    ReLU
    Conv2D, filters=48, kernel_size=[1, 1], input_shape=[17, 17, 168], stride=[1, 1], padding=0
    ReLU
    Conv2D, filters=12, kernel_size=[3, 3], input_shape=[17, 17, 48], stride=[1, 1], padding=1
SkipCat dim=1
SkipNet1
SkipNet2
    ReLU
    Conv2D, filters=48, kernel_size=[1, 1], input_shape=[17, 17, 180], stride=[1, 1], padding=0
    ReLU
    Conv2D, filters=12, kernel_size=[3, 3], input_shape=[17, 17, 48], stride=[1, 1], padding=1
SkipCat dim=1
SkipNet1
SkipNet2
    ReLU
    Conv2D, filters=48, kernel_size=[1, 1], input_shape=[17, 17, 192], stride=[1, 1], padding=0
    ReLU
    Conv2D, filters=12, kernel_size=[3, 3], input_shape=[17, 17, 48], stride=[1, 1], padding=1
SkipCat dim=1
SkipNet1
SkipNet2
    ReLU
    Conv2D, filters=48, kernel_size=[1, 1], input_shape=[17, 17, 204], stride=[1, 1], padding=0
    ReLU
    Conv2D, filters=12, kernel_size=[3, 3], input_shape=[17, 17, 48], stride=[1, 1], padding=1
SkipCat dim=1
SkipNet1
SkipNet2
    ReLU
    Conv2D, filters=48, kernel_size=[1, 1], input_shape=[17, 17, 216], stride=[1, 1], padding=0
    ReLU
    Conv2D, filters=12, kernel_size=[3, 3], input_shape=[17, 17, 48], stride=[1, 1], padding=1
SkipCat dim=1
SkipNet1
SkipNet2
    ReLU
    Conv2D, filters=48, kernel_size=[1, 1], input_shape=[17, 17, 228], stride=[1, 1], padding=0
    ReLU
    Conv2D, filters=12, kernel_size=[3, 3], input_shape=[17, 17, 48], stride=[1, 1], padding=1
SkipCat dim=1
SkipNet1
SkipNet2
    ReLU
    Conv2D, filters=48, kernel_size=[1, 1], input_shape=[17, 17, 240], stride=[1, 1], padding=0
    ReLU
    Conv2D, filters=12, kernel_size=[3, 3], input_shape=[17, 17, 48], stride=[1, 1], padding=1
SkipCat dim=1
SkipNet1
SkipNet2
    ReLU
    Conv2D, filters=48, kernel_size=[1, 1], input_shape=[17, 17, 252], stride=[1, 1], padding=0
    ReLU
    Conv2D, filters=12, kernel_size=[3, 3], input_shape=[17, 17, 48], stride=[1, 1], padding=1
SkipCat dim=1
SkipNet1
SkipNet2
    ReLU
    Conv2D, filters=48, kernel_size=[1, 1], input_shape=[17, 17, 264], stride=[1, 1], padding=0
    ReLU
    Conv2D, filters=12, kernel_size=[3, 3], input_shape=[17, 17, 48], stride=[1, 1], padding=1
SkipCat dim=1
SkipNet1
SkipNet2
    ReLU
    Conv2D, filters=48, kernel_size=[1, 1], input_shape=[17, 17, 276], stride=[1, 1], padding=0
    ReLU
    Conv2D, filters=12, kernel_size=[3, 3], input_shape=[17, 17, 48], stride=[1, 1], padding=1
SkipCat dim=1
SkipNet1
SkipNet2
    ReLU
    Conv2D, filters=48, kernel_size=[1, 1], input_shape=[17, 17, 288], stride=[1, 1], padding=0
    ReLU
    Conv2D, filters=12, kernel_size=[3, 3], input_shape=[17, 17, 48], stride=[1, 1], padding=1
SkipCat dim=1
ReLU
Conv2D, filters=150, kernel_size=[1, 1], input_shape=[17, 17, 300], stride=[1, 1], padding=0
AvgPool2D
SkipNet1
SkipNet2
    ReLU
    Conv2D, filters=48, kernel_size=[1, 1], input_shape=[9, 9, 150], stride=[1, 1], padding=0
    ReLU
    Conv2D, filters=12, kernel_size=[3, 3], input_shape=[9, 9, 48], stride=[1, 1], padding=1
SkipCat dim=1
SkipNet1
SkipNet2
    ReLU
    Conv2D, filters=48, kernel_size=[1, 1], input_shape=[9, 9, 162], stride=[1, 1], padding=0
    ReLU
    Conv2D, filters=12, kernel_size=[3, 3], input_shape=[9, 9, 48], stride=[1, 1], padding=1
SkipCat dim=1
SkipNet1
SkipNet2
    ReLU
    Conv2D, filters=48, kernel_size=[1, 1], input_shape=[9, 9, 174], stride=[1, 1], padding=0
    ReLU
    Conv2D, filters=12, kernel_size=[3, 3], input_shape=[9, 9, 48], stride=[1, 1], padding=1
SkipCat dim=1
SkipNet1
SkipNet2
    ReLU
    Conv2D, filters=48, kernel_size=[1, 1], input_shape=[9, 9, 186], stride=[1, 1], padding=0
    ReLU
    Conv2D, filters=12, kernel_size=[3, 3], input_shape=[9, 9, 48], stride=[1, 1], padding=1
SkipCat dim=1
SkipNet1
SkipNet2
    ReLU
    Conv2D, filters=48, kernel_size=[1, 1], input_shape=[9, 9, 198], stride=[1, 1], padding=0
    ReLU
    Conv2D, filters=12, kernel_size=[3, 3], input_shape=[9, 9, 48], stride=[1, 1], padding=1
SkipCat dim=1
SkipNet1
SkipNet2
    ReLU
    Conv2D, filters=48, kernel_size=[1, 1], input_shape=[9, 9, 210], stride=[1, 1], padding=0
    ReLU
    Conv2D, filters=12, kernel_size=[3, 3], input_shape=[9, 9, 48], stride=[1, 1], padding=1
SkipCat dim=1
SkipNet1
SkipNet2
    ReLU
    Conv2D, filters=48, kernel_size=[1, 1], input_shape=[9, 9, 222], stride=[1, 1], padding=0
    ReLU
    Conv2D, filters=12, kernel_size=[3, 3], input_shape=[9, 9, 48], stride=[1, 1], padding=1
SkipCat dim=1
SkipNet1
SkipNet2
    ReLU
    Conv2D, filters=48, kernel_size=[1, 1], input_shape=[9, 9, 234], stride=[1, 1], padding=0
    ReLU
    Conv2D, filters=12, kernel_size=[3, 3], input_shape=[9, 9, 48], stride=[1, 1], padding=1
SkipCat dim=1
SkipNet1
SkipNet2
    ReLU
    Conv2D, filters=48, kernel_size=[1, 1], input_shape=[9, 9, 246], stride=[1, 1], padding=0
    ReLU
    Conv2D, filters=12, kernel_size=[3, 3], input_shape=[9, 9, 48], stride=[1, 1], padding=1
SkipCat dim=1
SkipNet1
SkipNet2
    ReLU
    Conv2D, filters=48, kernel_size=[1, 1], input_shape=[9, 9, 258], stride=[1, 1], padding=0
    ReLU
    Conv2D, filters=12, kernel_size=[3, 3], input_shape=[9, 9, 48], stride=[1, 1], padding=1
SkipCat dim=1
SkipNet1
SkipNet2
    ReLU
    Conv2D, filters=48, kernel_size=[1, 1], input_shape=[9, 9, 270], stride=[1, 1], padding=0
    ReLU
    Conv2D, filters=12, kernel_size=[3, 3], input_shape=[9, 9, 48], stride=[1, 1], padding=1
SkipCat dim=1
SkipNet1
SkipNet2
    ReLU
    Conv2D, filters=48, kernel_size=[1, 1], input_shape=[9, 9, 282], stride=[1, 1], padding=0
    ReLU
    Conv2D, filters=12, kernel_size=[3, 3], input_shape=[9, 9, 48], stride=[1, 1], padding=1
SkipCat dim=1
SkipNet1
SkipNet2
    ReLU
    Conv2D, filters=48, kernel_size=[1, 1], input_shape=[9, 9, 294], stride=[1, 1], padding=0
    ReLU
    Conv2D, filters=12, kernel_size=[3, 3], input_shape=[9, 9, 48], stride=[1, 1], padding=1
SkipCat dim=1
SkipNet1
SkipNet2
    ReLU
    Conv2D, filters=48, kernel_size=[1, 1], input_shape=[9, 9, 306], stride=[1, 1], padding=0
    ReLU
    Conv2D, filters=12, kernel_size=[3, 3], input_shape=[9, 9, 48], stride=[1, 1], padding=1
SkipCat dim=1
SkipNet1
SkipNet2
    ReLU
    Conv2D, filters=48, kernel_size=[1, 1], input_shape=[9, 9, 318], stride=[1, 1], padding=0
    ReLU
    Conv2D, filters=12, kernel_size=[3, 3], input_shape=[9, 9, 48], stride=[1, 1], padding=1
SkipCat dim=1
SkipNet1
SkipNet2
    ReLU
    Conv2D, filters=48, kernel_size=[1, 1], input_shape=[9, 9, 330], stride=[1, 1], padding=0
    ReLU
    Conv2D, filters=12, kernel_size=[3, 3], input_shape=[9, 9, 48], stride=[1, 1], padding=1
SkipCat dim=1
ReLU
AvgPool2D
CorrelateAll only_train=False
Linear out=10
\end{Verbatim}
\end{strip}
\quad
\clearpage

\fi

\end{document}